\documentclass[pdflatex,sn-mathphys-num]{sn-jnl}

\usepackage{graphicx}%
\usepackage{multirow}%
\usepackage{amsmath,amssymb,amsfonts}%
\usepackage{amsthm}%
\usepackage{mathrsfs}%
\usepackage[title]{appendix}%
\usepackage{xcolor}%
\usepackage{textcomp}%
\usepackage{manyfoot}%
\usepackage{booktabs}%
\usepackage{algorithm}%
\usepackage{algorithmicx}%
\usepackage{algpseudocode}%
\usepackage{xcolor}
\usepackage[normalem]{ulem}

\usepackage{algorithm}
\usepackage{algpseudocode}

\theoremstyle{thmstyleone}%
\theoremstyle{thmstyletwo}%

\theoremstyle{thmstylethree}%

\begin{document}

\title{A discrete-continuous-time stochastic framework for attention-deficit/hyperactivity disorder-related symptom modeling and pharmacological intervention via methylphenidate}


\author*[1]{\fnm{Oana-Maria-Cătălina} \sur{Crişan}}\email{oana.maria.crisan@stud.ubbcluj.ro}

\affil*[1]{\orgdiv{Faculty of Mathematics and Computer Science}, \orgname{Babeș-Bolyai University}, \orgaddress{\street{Str. Mihail Kogălniceanu, nr. 1}, \city{Cluj-Napoca}, \postcode{400084}, \country{Romania}}}

\abstract{This paper presents an original mathematical and computational framework for modeling the efficacy of methylphenidate (MPH) as an optimal control policy acting on the nonlinear dynamics of attention-deficit/hyperactivity disorder (ADHD). Neural processes in ADHD are conceptualized as stochastic processes with trajectories that inherently converge towards stable, yet dysfunctional patterns associated with clinical symptoms. The pharmacological intervention is mathematically formalized as an external control signal that reassesses the system’s drift and redirects these trajectories towards neurotypical (NT) regimes. To bridge these theoretical stochastic models with empirical clinical data, latent representations are inferred using a novel deterministic hybrid deep learning architecture, combining a one-dimensional convolutional neural network (1D-CNN) and a Vision Transformer (ViT), trained on preprocessed functional magnetic resonance imaging (fMRI) data.
}


\maketitle

\section{Introduction}

\hspace{1em} ADHD is classified as a neurodevelopmental condition characterized by poor attention management and/or infiltratory patterns of hyperactivity and/or impulsivity \cite{APA2013}, which profoundly interfere with daily executive functioning. There are three different presentations of ADHD: combined presentation (ADHD-C), predominantly inattentive presentation (ADHD-I), predominantly hyperactive/impulsive presentation (ADHD-HI) \cite{SAMHSA2016}. ADHD is traditionally known to be preponderantly diagnosed in the early ages of life, but over the past decade there has been a growing recognition of its persistence into adulthood. Global prevalence estimates range from 5\% to 10\% in children \cite{Cai2026, Miranda2026}, and longitudinal data reveal that the preponderance of individuals that continue to experience clinically relevant symptoms later in the adulthood revolves around 70\% \cite{Miranda2026}. Consequently, the range among the gender based diagnosis display major variations \cite{Yuan2024}, since males tend to manifest externalizing behaviors, whereas females show a tendency to internalize their symptoms, which noticeably leads to underdiagnosis in female population \cite{Poletto2025}. A mathematical description of ADHD can conceptualize neural processes as stochastic processes whose trajectories inherently converge toward stable yet dysfunctional patterns associated with clinical symptoms. More precisely, the nonlinear dynamics of the symptoms are formalized as a system of stochastic differential equations (SDEs). In this framework, the pharmacological intervention acts as an external control signal that modifies the system's drift and redirects trajectories toward neurotypical (NT) regimes. The model explicitly encodes four distinct behavioral axes of ADHD: attention decay, motivation loss, hyperactivity and impulsivity, that are translated and utilized as quantifiable, decoupled scalar parameters: $b_{\text{eff}}$, $l_{\text{eff}}$, $H_{\text{eff}}$, and $k_{\text{disc}}$.

\par Methylphenidate (MPH) is the preferred approach through the pharmaceutical field in terms of handling ADHD across the lifespan \cite{Miranda2026}. From a hormonal and synaptic point of view, MPH acts by inhibiting the dopamine (DAT) and norepinephrine (NET) transporters, thus blocking the reuptake and conserving the synaptic availability of these catecholamines \cite{Cai2026, Miranda2026}. It acts as a strong external inhibitor that down, as it regulates the maximum rate of reuptake ($V_{\text{max}}$) of dopamine in the synaptic cleft \cite{Veronneau2022}. The work focuses on extended-release (ER) MPH formulations, which use the OROS osmotic delivery system to moderate fluctuations \cite{Gomeni2020, Miranda2026}. The studies most relevant to methylphenidate modeling include \cite{Gomeni2020}, regarding the bi-exponential approximation of the 12-hour dissolution profile characteristic of OROS ER systems. Additionally, the work of \cite{Veronneau2022} is essential for parameterizing the mechanism of action via Michaelis-Menten enzymatic kinetics. The study by \cite{Swanson1999} is used to model the phenomenon of tachyphylaxis (or acute receptor tolerance), while the approach taken by \cite{Mortensen2008} addresses the limitations of classical ordinary differential equations (ODEs) in pharmacokinetics and supports the use of SDEs. \cite{Park2024} is relevant for establishing the basis of the stochastic optimal control problem. In this context, the paper presents an original mathematical and computational framework designed to model the efficacy of MPH as an optimal control policy regarding disease dynamics. The approach models clinical treatment into a deterministic optimization problem involving stochastic processes, implementing a cost functional ($\mathcal{J}$) that dynamically calculates the precise optimal dose ($\mathcal{D}^*$) required to stabilize maladaptive behavioral trajectories.

\par The observational input consists of raw fMRI data-specifically, resting-state functional sequences in the form of standardized continuous arrays (formatted as .nii.gz volumes). These are sourced from open-access collections hosted by the Neuroimaging Informatics Tools and Resources Clearinghouse (NITRC), integrating data from the ADHD-200 repository (preprocessed via the Athena pipeline) and the Autism Brain Imaging Data Exchange (ABIDE-I). The data are parcellated into exactly 400 regions of interest (ROIs) using the Schaefer atlas, normalized via Z-scores, and regularized using Gaussian noise. These multivariate time series are subsequently used to train a hybrid deep learning architecture (combining a 1D-CNN and a Vision Transformer) to extract temporal and spatial features and infer global latent representations.

\par The central mathematical question concerns formulating the psychostimulant intervention as a formal stochastic optimal control (SOC) problem to counteract the system's dysfunctional drifts. From a mathematical perspective, the study models patients' neurobiological processes using a system of stochastic differential equations (SDEs) that incorporates biological noise via Brownian motion. The approach entails a two-step computational pipeline: first, the latent representation is deterministically mapped to define the physical constraints and parameters of unmedicated baseline trajectories. Subsequently, the optimal control problem is solved using a numerical solver (Brent's method) to minimize a metabolically constrained cost functional, correlating the ER MPH pharmacokinetic profile (acting as the control policy) with the aim of stabilizing the patient's decisions and behaviors toward neurotypical targets. Within the proposed framework, the 128-dimensional global latent representation, $\mathbf{Z}_{\text{global}}$, is explicitly partitioned into four contiguous 32-dimensional segments to parameterize the symptom dynamics. This deterministic slicing acts as a deliberate architectural constraint, or inductive bias, mapping each latent chunk via scaled sigmoid functions directly to a specific clinical parameter: $b_{\text{eff}}$ for attention decay, $l_{\text{eff}}$ for motivation loss, $H_{\text{eff}}$ for hyperactivity amplitude, and $k_{\text{disc}}$ for temporal discounting. Rather than allowing the neural network to learn an arbitrary, entangled representation, the regularized optimization loop-specifically through the integration of the continuous symptom penalty-forces the hybrid 1D-CNN and ViT encoder to orthogonally route the extracted neuroimaging features into these predefined compartments. Consequently, this design ensures that the latent space avoids stochastic ambiguity and remains strictly interpretable, providing a biophysically grounded parameterization for the subsequent stochastic differential equations.
\begin{equation}
  \left\{
    \begin{aligned}
    A(t) &= 0.2 + 0.8 \cdot \exp(-b \cdot t) \\
    M(t) &= 0.2 + 0.8 \cdot \exp(-l \cdot t) \\
	I(t) &= 0.8 - 0.6 \cdot \exp(-k \cdot t) \\
	dH_t &= \theta (\mu(t) - H_{t-1}) \, dt + \sigma dW_t,
	\end{aligned}
  \right.
\end{equation}
where $\mu(t) = 0.2 + 0.8 \cdot \exp(-h \cdot t)$, $A(t) := A_{\text{ND}}(t)$, $M(t) := M_{\text{ND}}(t)$ and $I(t) := I_{\text{ND}}(t)$ denote the unmedicated baseline trajectories for inattention, motivation deficit and impulsivity, respectively. The parameters deterministically mapped from the latent space are defined as follows: $b := b_{\text{eff}}^{A}$ and $l := l_{\text{eff}}^{M}$ represent the patient-specific exponential decay rates for cognitive resources and executive effort, while $k := k_{\text{disc}}$ governs the temporal discounting speed driving the pathological growth of impulsivity. For the hyperactivity dimension, $H_t$ is formulated as an Ornstein-Uhlenbeck process reverting toward a periodic target mean $\mu(t) := \mu_{\text{med}}(t)$ under pharmacological control, where $h := H_{\text{eff}}$ specifies the amplitude of the cyclical motor restlessness. Furthermore, $\theta$ defines the mean-reversion strength, and $\sigma dW_t$ encapsulates the inherent physiological volatility via a standard Wiener process increment.

\section{Stochastic modeling in computational psychiatry}\label{sec1}

\subsection{Regularized empirical risk minimization}

Unlike traditional probabilistic generative models that rely on variational approximations or the evidence lower bound (ELBO) \cite{Park2024}, our framework adopts a strict, deterministic approach to align the latent dynamics with clinical observations. To solve the control-affine problem and extract the global latent representation \(\mathbf{Z}_{\text{global}}\), we employ a deterministic neural network optimized via Regularized Empirical Risk Minimization.

The primary objective of the network is formulated as a strict supervised binary classification task, distinguishing between ND and NT subjects. We achieve this through MLE, utilizing the binary cross-entropy (BCE) loss function \cite{Ciampiconi2026}. 

To ensure that the learned representations remain biophysically meaningful, we augment the primary classification objective with a custom regularization term \\ \cite{Ciampiconi2026}. This regularization, termed the symptom penalty, is conceptually derived from the optimal control energy required to modulate the SDEs governing the symptoms \cite{Park2024}. By penalizing unrealistic ND generated trajectories, the total loss function balances diagnostic accuracy with the energetic cost of the pharmacological control policy. The network is therefore optimized using the following hybrid loss function, weighted by a hyperparameter \(\lambda = 0.1\), that was empirically determined through hyperparameter tuning, in order to ensure the penalty term restrains extreme maladaptive divergence without overshadowing the primary classification gradient:
\begin{equation}
	T_{\text{loss}} = \text{BCE} + \left( \lambda \cdot S_{\text{pnty}} \right),
\end{equation}
where $T_{\text{loss}}$ represents the total loss and $S_{\text{pnty}}$ is the symptom penalty term.

By optimizing this regularized functional, the neural network deterministically maps the high-dimensional fMRI sequences to the exact clinical SDE parameters, entirely bypassing the computational bottlenecks of stochastic latent sampling.

\subsection{Data preprocessing and feature extraction through 1D-CNN}

Raw fMRI data is highly complex and heavily polluted with physiological noise. This subsection outlines the rigorous preprocessing pipeline and the deployment of a 1D-CNN. Operating as a dedicated temporal feature extractor, the 1D-CNN identifies critical, short-term hemodynamic micro-fluctuations, effectively stabilizing the time-series before it reaches the deeper attention mechanisms of the architecture.

To ensure numerical stability and mitigate the risk of overfitting, the input data requires strict preprocessing. Initially, the model receives fMRI sequences parcellated into exactly 400 ROI, using the Schaefer atlas of 400 brain parcellations. This multivariate time-series data is standardized using Z-score normalization and regularized via the addition of Gaussian noise, enhancing the model's generalization capabilities across heterogeneous patient scans \cite{Jalal2025}.

To process these sequences, the first stage of our architecture utilizes the 1D-CNN applied along the temporal axis. Benefiting from a strong local inductive bias, the 1D-CNN acts as a dedicated feature extractor that efficiently captures short-term temporal micro-fluctuations, such as the hemodynamic response, while simultaneously suppressing physiological scanner noise \cite{Lamprou2026}.

\subsection{Global connectivity through ViT and temporal pooling}

Once local temporal features are extracted, the model must comprehend the brain's global functional connectome. This subsection introduces the ViT mechanism adapted for time-series data. By applying self-attention across the sequence and utilizing average temporal pooling, the network successfully compresses the dynamic brain states into a single, highly dense, and deterministic global latent representation vector.

These extracted local features are subsequently combined with a Learnable Positional Encoding to explicitly preserve the temporal dynamics and sequential order of the brain states. The encoded tokens are passed through a transformer encoder, adopting a ViT methodology tailored for time-series. This module models the long-range global connectivity of the brain network through the scaled dot-product self-attention mechanism:
\begin{equation}
	\text{self}\text{\_att}(\mathbf{Q}, \mathbf{K}, \mathbf{V}) = \text{softmax}\left(\frac{\mathbf{Q}\mathbf{K}^\top}{\sqrt{d_k}}\right)\mathbf{V}
\end{equation}
where \(\mathbf{Q}\), \(\mathbf{K}\), and \(\mathbf{V}\) represent the query, key, and value matrices, and \(d_k\) is the scaling dimension \cite{Jalal2025}.

The critical final step requires transitioning from a sequence of temporal tokens to a fixed-size representation. We apply an average temporal pooling layer directly across the entire time dimension of the transformer encoder's output \cite{Lamprou2026, Jalal2025}. This operation condenses the functional connectome into a single, deterministic global latent representation, denoted as \(\mathbf{Z}_{\text{global}}\). By entirely avoiding stochastic sampling, \(\mathbf{Z}_{\text{global}}\) serves as a precise, dense vector that subsequently parameterizes the deterministic baseline symptom dynamics and the required pharmacological control policy.

\section{Pharmacokinetic and pharmacodynamic models}

\subsection{Pharmacokinetics and extended-release formulation}

The delivery mechanism of a drug defines its clinical efficacy. In this paper, we mathematically model the OROS delivery system used in modern ER MPH formulations. By employing a double Weibull bi-exponential function, we accurately simulate the initial immediate-release peak and the sustained plateau, crafting a deterministic control signal that perfectly respects physiological boundaries without introducing unrealistic stochastic noise.

\bigbreak

\par Generally, the development of an orally administered drug concentration through the circulatory system is governed by an ordinary differential equation using a one-compartment model. The difference between the absorption and elimination rates directly determines the drift ratio, that is:
\begin{equation}\label{eq:odePK}
	\dfrac{d\mathbf{C}_t}{dt} = f(t) - k \mathbf{C}_t,
\end{equation}
\noindent where \(\mathbf{C}_t\) embodies the drug concentration at time \(t\) \(\left( t \in [0, T], T < \infty \right)\), \(f(t)\) represents the absorption rate and the constant $k > 0$ is referred to as the elimination rate \cite{Mortensen2008, Gomeni2020}.

\par MPH has remained the preferred approach through the pharmaceutical field in terms of handling ADHD across the lifespan. At first, the treatment protocols were heavily grounded in immediate-release (IR) compositions, which brought with them substantial PK boundaries, particularly its short half-life, leading to an unsuitable effect span, ranging solely between two to three hours, demanding several daily dosages; in consequence, the plasma concentrations began to extensively fluctuate, damaging the hormonal system \cite{Miranda2026}. 

\par Notwithstanding, provided these aforementioned limitations, there were developed extended-release (ER) MPH formulations that use the advanced engineering delivery system, named Osmotic-Controlled Release Oral Delivery System (OROS)\footnote{According to \cite{Miranda2026}, the OROS delivery system utilizes an osmotic pump mechanism to achieve controlled, continuous drug release over an extended period. Following oral administration, gastrointestinal fluids enter the tablet through a semipermeable membrane, generating osmotic pressure that gradually forces methylphenidate through a precision laser-drilled orifice. ER MPH also includes an immediate-release outer layer, designed to provide an initial rise in plasma concentration shortly after ingestion.}, such as OROS MPH, commercialized under the label Concerta by Janssen Pharmaceuticals. Functioning under the OROS technology, 22\% of the dose is immediately released, whereas 78\% being gradually delivered over ~12 hours, through the laser-speared orifice, tempering the overwhelming peaks and troughs associated with IR compounds. In this way, ER formulations ensure consistent and sustained symptom control throughout the day \cite{Gomeni2020, Miranda2026}. In this paper, we are focusing on the latter compositions.

\par Since OROS MPH uses an osmotic pump to delay and extend the release throughout the day, it is not instantly absorbed. Consequently, the strict first-order assumptions of classic ordinary differential equations (ODEs) fall short in capturing the sustained, biphasic plateau characteristic of the OROS delivery mechanism. For this matter, we define the double Weibull function, that captures both initially IR and ER processes, allowing us to accurately reproduce this ER profile, as follows:
\begin{equation}\label{eq:doubWeibull}
	\varphi(t) = \mathcal{D}\left[ w^{f} \exp \left(\left( -\dfrac{t}{\delta_{\text{IR}}} \right)^{\sigma\Gamma_{\text{IR}}} \right) + (1 - w^{f})
	\exp \left(\left(-\dfrac{t}{\delta_{\text{ER}}} \right)^{\sigma\Gamma_{\text{ER}}} \right) \right],
\end{equation}
\noindent where \(\varphi(t)\) is the generic input function. In this paper's context, the weights \(w^{f} = 0.22\) and \((1 - w^{f}) = 0.78\) correspond to the dose fractions immediately and extensively released, respectively. The time required to absorb 63.2\% of the first and second fractions is expressed by \(\delta_{\text{IR}}\) and \(\delta_{\text{ER}}\), respectively. Furthermore, \(\sigma\Gamma_{\text{IR}}\) is the parameter that determines the shape of a sigmoidal curve\footnote{A sigmoid curve is an S-shaped mathematical function that maps a set of values over a finite interval; it describes a process that starts slowly, accelerates suddenly, and then plateaus at a maximum saturation \cite{Bishop2006}. In absorption modeling, the shape of this curve is determined by the sigmoidicity factor \cite{Gomeni2020}.}, often named as sigmoidicity factor, in the IR process, whereas \(\sigma\Gamma_{\text{ER}}\) stands for the sigmoidicity factor corresponding to the ER process. Moreover, \(\mathcal{D}\) represents the optimal scalar dose identified by our numerical solver, normalized against the standard 54 mg adult benchmark.

\par In order to apply this to our computational framework, the specific PK profile mapped by our algorithm is formulated as a bi-exponential approximation inspired by the ER models \cite{Gomeni2020}, as follows:
\begin{equation}\label{eq:specificPK}
	\text{PK}(t) = \left( 0.95 \left( 1 - \exp\left(-\frac{t}{0.8}\right) \right) \exp\left(-\frac{t}{16}\right) \right) \frac{\mathcal{D}}{54}.
\end{equation}
The constants within this formulation were explicitly calibrated to approximate the clinical 12-hour OROS dissolution profile. The 0.8 parameter governs the rapid initial absorption phase, i.e., the IR overcoat, effectively simulating the swift dopamine reuptake blockade needed in the early morning. Conversely, the 16 parameter mathematically stretches the exponential tail, parameterizing the slow, extended clearance driven by the osmotic pump.

\par Grounding on the general ODE \ref{eq:odePK}, the absorption rate \(f(t)\) represents the derivative of the double Weibull function \ref{eq:doubWeibull}, i.e.,
\begin{equation}
	f(t) = \dfrac{d\varphi}{dt}.
\end{equation}

\par In terms of hormonal activity, MPH operates by inhibiting the dopamine (DAT) and norepinephrine (NET) transporters, thus blocking reuptake and conserving the synaptic availability of these catecholamines. In contrast with amphetamines, MPH has a considerably safer PK profile, since it does not forthrightly advocate presynaptic neurotransmitter release \cite{Cai2026, Miranda2026}. Therefore, we observe the incapacity of simple ODEs to model the whole organic dynamics in neurodivergent brains, given that they struggle to reach the noisy underlying fluctuations in organic processes. However, although it may feel counterintuitive, it has been proven that adding stochastic noise straight to the concentration variation is a bad practice, because the Wiener process can generate drops that may lead to negative concentrations \cite{Mortensen2008}, which is physiologically impossible. 

\par Thus in order to resolve this limitation while preserving computational stability, we deliberately avoid injecting high-frequency stochastic noise directly into the PK rates. Instead, the specific bi-exponential \(\text{PK}(t)\) profile is implemented as a strictly deterministic external control signal, \(\mathbf{u}_t\). The inherent biological and metabolic uncertainties are mathematically abstracted and lumped into the general diffusion term (\(\sigma dW_t\)) of the subsequent symptom SDEs. This strategic biophysical abstraction not only prevents the generation of impossible negative drug concentrations, but also guarantees a smooth, continuous cost surface, ensuring the reliable and unique convergence of Brent's Method \cite{Brent1973} during the optimal dose computational phase.

\par Furthermore, to accurately translate this deterministic \(\text{PK}(t)\) profile into a functional computational intervention, the control policy \(\mathbf{u}_t\) is not applied as a static, unbounded additive term. Instead, our algorithmic implementation dictates that the medication acts proportionally to the patient's immediate symptom deficit. Mathematically, the effective restorative force exerted on any given symptom trajectory \(S(t)\) is formulated as:
\begin{equation}
	\mathbf{u}_t^{\text{eff}} = \text{PK}(t) \cdot (1 - S_{\text{ND}}(t))
\end{equation}
We implemented this specific formulation because it introduces a crucial biological safeguard into the optimization framework. By multiplying the PK concentration by the distance to the NT target \((1.0 - S_{\text{ND}}(t))\), the intervention respects the physiological bounds of the patient. If the simulated state is already close to the healthy $1.0$ baseline, the multiplier approaches zero, naturally tapering the drug's influence and preventing substance overshoots or simulated toxic saturation. Conversely, during periods of severe symptom degradation (e.g., approaching the $0.2$ baseline), the restorative force is dynamically maximized. This proportional biophysical safeguard is computationally implemented in Python using vectorized operations to ensure numerical efficiency across the entire 12-hour simulation array, as follows:

\subsection{Pharmacodynamics and tachyphylaxis}

This subsection translates the PK concentration into a dynamic PD response, modeling the inhibition of the DAT via Michaelis-Menten kinetics. Furthermore, we mathematically integrate the phenomenon of tachyphylaxis, ensuring our model accurately replicates the inevitable late-day symptom resurgence observed in clinical settings.

\par Within the mathematical architecture of our SOC framework, this deterministic function \(\text{PK}(t)\) transcends its biological definition to physically instantiate the continuous external control policy, denoted as \(\mathbf{u}_t\) \cite{Park2024}. Rather than acting as a static multiplier, the temporal evolution of \(\mathbf{u}_t\) operates directly against the baseline drift \(\mathbf{f}(\mathbf{X}_t)\). The optimal control policy forcefully steers the probability distribution of the latent brain state trajectories, continuously minimizing the divergence between the ND dynamics and the healthy NT target baseline.

\par At the neurobiological level, the PD of MPH are governed by its profound affinity for the DAT in the striatal and prefrontal networks \cite{Volkow2011, Veronneau2022}. In our model, this mechanism of action is parameterized through classic Michaelis-Menten enzymatic kinetics. Specifically, the control policy \(\mathbf{u}_t\) acts as a potent external inhibitor that mathematically down-regulates the maximal reuptake rate (\(V_{\text{max}}\)) of dopamine in the synaptic cleft. We formalize the dopamine dynamics using the following equation \cite{Veronneau2022}:
\begin{equation}
	\frac{d\textbf{C}_{\text{DA}}(t)}{dt} = I_{\text{tonic}} + I_{\text{phasic}}(t) - \frac{V_{\text{max}}\textbf{C}_{\text{DA}}(t)}{k_m + \textbf{C}_{\text{DA}}(t)} - k_{\text{rem}}\textbf{C}_{\text{DA}}(t),
\end{equation}
\noindent where \(C_{\text{DA}}(t)\) is the synaptic dopamine concentration, \(I_{\text{tonic}}\) and \(I_{\text{phasic}}(t)\) represent the sustained and transient dopamine release, \(k_{\text{rem}}\) denotes the linear removal rate, and \(k_m\) is the apparent Michaelis-Menten constant\footnote{According to \cite{Veronneau2022}, the apparent Michaelis-Menten constant characterizes the saturable nature of the dopamine recapture process by DATs, defining the exact substrate concentration at which the reuptake rate reaches half of its maximum capacity (\(V_{\textbf{max}}\)).}. Under unmedicated conditions, an elevated \(V_{\text{max}}\) leads to the rapid clearance of extracellular dopamine. By implicitly controlling \(V_{\text{max}}\) through the OROS preparation, the formulation ensures a robust accumulation of synaptic dopamine.

\par A critical physiological hurdle in sustained psychostimulant therapy is the rapid onset of acute tolerance, known clinically as tachyphylaxis \cite{Swanson1999}. PD models reveal that when extracellular dopamine levels remain static, postsynaptic receptors rapidly desensitize. To mathematically model this limitation, a virtual antagonist \(\textbf{C}_{\text{ant}}\) that dampens the therapeutic effect is introduced \cite{Swanson1999}:
\begin{equation}
	E(t) = E_0 + S \cdot \frac{\textbf{C}_e(t)}{1 + \frac{\textbf{C}_{\text{ant}}(t)}{\textbf{C}_{\text{ant50}}}}
\end{equation}
\noindent where \(E_0\) is the baseline effect, \(S\) is the slope of the linear concentration-effect relationship, \(\textbf{C}_e(t)\) is the concentration at the effect site, \(\textbf{C}_{\text{ant}}(t)\) is the antagonist concentration, and \(\textbf{C}_{\text{ant50}}\) reduces the agonist effect by 50\%. 

\par While an ideal OROS formulation theoretically aims for a continuously ascending profile to perfectly offset tachyphylaxis \cite{Swanson1999}, our parameterized bi-exponential approximation captures a more realistic, conservative biological degradation of the intervention. The interplay of the $0.8$ and $16$ Weibull parameters creates a rapid initial onset followed by a prolonged, slow-decaying exponential tail \cite{Gomeni2020}. Although this effectively extends the therapeutic window significantly compared to IR formulations, the inevitable gradual decline in plasma concentration after the early peak (typically observed around 3 to 4 hours post-ingestion) leads to a natural, mathematically simulated tapering of synaptic efficacy.
Consequently, the optimal control solver must dynamically account for this gradual loss of \(V_{\text{max}}\) blockade, preventing the indefinite, unrealistic suppression of symptoms and providing a highly accurate simulation of the patient's late-day clinical factuality.

\section{Stochastic dynamics of attention deficit/hyperactivity disorder symptoms}

Having established the direct mapping from the global latent representation \(\mathbf{Z}_{\text{global}}\) to the clinical parameters, we now formalize the continuous-time evolution of the core ADHD symptoms. Rather than relying on static behavioral snapshots, we model the symptom dynamics as controlled SDEs, where the drift parameters are represented by the patient-specific features extracted from the latent space \cite{Park2024, Adamou2026}.

\subsection{Inattention as an exponential decay}

Sustained focus is modeled as a swiftly depleting cognitive resource. This subsection formalizes inattention as a stochastic exponential decay process, bounded by a strict clinical floor. The decay rate is dynamically extracted from the patient's latent representation, allowing the control policy to act as a proportional restorative force against cognitive degradation.

Sustained attention in the ND brain is characterized by a rapid depletion of cognitive resources. We formalize inattention, denoted as \(A_t\), as a stochastic exponential decay process pulling the cognitive state toward an dysfunctional baseline \cite{Adamou2026}. Unlike standard unbounded SDEs, we mathematically constrained this process by an explicit clinical inductive bias. 

The baseline trajectories follow the analytical solution:
\begin{equation}
	A_{\text{ND}}(t) = 0.2 + 0.8 \cdot \exp(-b_{\text{eff}}^{A} \cdot t)
\end{equation}
In this formulation, the constant $0.2$ acts as the ND asymptote. We explicitly fix this minimum to reserve \(20\%\) residual functioning, avoiding an absolute zero, which would be biologically unrealistic. The parameter \(b_{\text{eff}}^{\text{A}} \in [0.10, 0.50]\), deterministically extracted from the latent segment \(l_{\text{vec}}[0:32]\), defines the specific learning rate of decay. It dictates how fats the patient's cognitive resources deplete from the functional $1.0$ state down to the $0.2$ floor. 


\subsection{Motivation deficit and effort expenditure}

Driven by underlying dopaminergic deficits, a patient's capacity to maintain goal-directed effort deteriorates rapidly. Here, we model motivation deficit using a decoupled, bounded stochastic exponential decay. By isolating its decay rate from inattention, the system captures the independent, patient-specific exhaustion of executive effort.

Similarly, the capacity to maintain goal-directed effort and motivation, \(M_t\), decays fast due to the underlying hypodopaminergic state \cite{Cai2026}. The deterioration of motivation is formalized using an analogous bounded stochastic exponential decay, but is governed by its own independent dynamics extracted from the latent space.

The baseline trajectory is given by:
\begin{equation}
	M_{\text{ND}}(t) = 0.2 + 0.8 \cdot \exp(-l_{\text{eff}}^{M} \cdot t)
\end{equation}
While constrained by the same rigid biological floor of $0.2$, the speed of this decay is independently governed by \(l_{\text{eff}}^{M} \in [0.05, 0.40]\). This parameter, extracted from \(l_{\text{vec}}[32:64]\), ensures that motivation and attention can degrade at patient-specific, decoupled rates.

\subsection{Hyperactivity as a sinusoidal Ornstein-Uhlenbeck process}

Hyperactivity is inherently chaotic, defined by motor restlessness that irregularly fluctuates, reverting to a dysfunctional baseline. This subsection models hyperactivity using an Ornstein-Uhlenbeck process guided by a periodic sinusoidal target; integrated through the Euler-Maruyama method and constrained by biological hard-clipping, this SDE accurately simulates the cyclical bursts of agitation and exhaustion observed in ND patients.

Since traditional models fail to capture the chaotic nature of motor restlessness, we formalize hyperactivity, \(H_t\), as a mean-reverting Ornstein-Uhlenbeck process that tracks a periodic target. The target mean \(\mu(t)\) is modulated to simulate macroscopic behavioral fluctuations:
\begin{equation}
	\mu(t) = 0.35 + H_{\text{eff}} \cdot \sin(2\pi \cdot 0.4 \cdot t).
\end{equation}
The frequency constant of $0.4$ was calibrated empirically to generate a full oscillatory cycle approximately every $2.5$ hours, realistically simulating the cyclical periods of intense restlessness followed by temporary exhaustion observed in unmedicated patients over a 12-hour clinical window . Here, \(H_{\text{eff}} \in [0.15, 0.60]\), extracted from \(l_{\text{vec}}[64:96]\), represents the amplitude of the hyperactivity. Unlike attention, which decays steadily, hyperactivity requires this parameter to dictate the intensity of periodic motor bursts. 

The numerical evolution is computed using the Euler-Maruyama method:
\begin{equation}
	dH_t = \theta (\mu_{\text{med}}(t) - H_{t-1}) \, dt + \sigma dW_t.
\end{equation}
The reversion strength is fixed at \(\theta = 2.0\), calibrated experimentally to balance structural mean-reversion with physiological noise. In order to enforce biological realism, we implement an explicit hard-clipping function \(S_{\text{final}} = \text{clip}(S, 0, 1.05)\). The $1.05$ upper bound provides a \(5\%\) tolerance for natural physiological overshoots or hyper-focus states, preventing numerical instability and impossible negative states within the integrator.

\subsection{Impulsivity and temporal discounting}

Impulsivity \(I_t\) is governed by the temporal discounting parameter \(k_{\text{disc}} \in [0.05, 0.45]\) extracted from \(l_{\text{vec}}[96:128]\). In contrast to the decay observed in attention, impulsivity exhibits a defective uncontrolled growth:
\begin{equation}
	I_{\text{ND}}(t) = 0.8 - 0.6 \cdot \exp(-k_{\text{disc}} \cdot t)
\end{equation}
In this formulation, impulsivity worsens over time, starting from a functional baseline of $0.2$ and saturating at $0.8$ as decision fatigue sets in. The parameter \(k_{\text{disc}}\) dictates the speed of this temporal devaluation. The pharmacological control policy \(u_t\) operates by suppressing this growth, pulling the trajectory toward a rigorous healthy threshold of $0.1$ to simulate stabilized decision-making processes under intervention \cite{Park2024}.

\section{Psychostimulants as a stochastic optimal control problem}

Unlike traditional probabilistic generative models, our framework rejects the use of variational inference and the ELBO. Instead, we formulate the pharmacological intervention through a two-step computational pipeline: deterministic representation learning for symptom parameterization and independent numerical dose optimization.

In the first step, our hybrid 1D-CNN + ViT architecture extracts the global latent representation \(\mathbf{Z}_{\text{global}}\) from the preprocessed fMRI sequences using average temporal pooling \cite{Jalal2025}. Rather than relying on stochastic latent sampling, we employed the network to act as a strict supervised binary classifier optimized via MLE using BCE combined with an innovational symptom penalty. The clinical parameters defining the unmedicated ADHD state
\begin{equation*}
	(b_{\text{eff}}, l_{\text{eff}}, H_{\text{eff}}, k_{\text{disc}})
\end{equation*}
are mapped deterministically directly from the latent vector ViT \(\mathbf{Z}_{\text{global}}\), where we used scaled sigmoid functions. These parameters dictate the deterministic hard-boundaries of the symptom equations. Consequentially, the hyperactivity dimension is governed by a mean-reverting Ornstein-Uhlenbeck process \cite{Park2024, Uhlenbeck1930}, as follows:
\begin{equation}
	dH_t = \theta (\mu_{\text{ND}}(t) - H_t) dt + \sigma dW_t
\end{equation}
which is solved via the Euler-Maruyama numerical integration method, ensuring the simulated symptom arrays remain biologically realistic within a strict \([0, 1.05]\) clipped boundary \cite{Kloeden2013, Mortensen2008}.

Once the unmedicated ND baseline is established, we aim to find the exact intervention required to steer the precise trajectories toward the NT baseline. In our framework, the continuous control policy \(u_t\) is physically instantiated by the ER PK profile of OROS MPH, modeled accurately using a bi-exponential Weibull function \cite{Gomeni2020}.

Rather than simulating a fixed dosage or utilizing amortized control networks, we treat dose selection as a constrained scalar optimization problem. The optimal patient-specific dose \(\mathcal{D}^*\) is found by minimizing the cost functional \(\mathcal{J}(\mathcal{D})\):
\begin{equation}
	\mathcal{J}(\mathcal{D}) = \text{MSE}_{\text{tracking}} + \lambda_{\text{metabolic}} \cdot \mathcal{D}
\end{equation}
where \(\text{MSE}\text{\_tracking}\) represents the mean squared error (MSE) between the medicated trajectories and the NT targets \([1.0, 1.0, 0.2, 0.1]\). To prevent excessive dosing in underaged patients, we introduce an age-dependent metabolic cost penalty, defined below.
\begin{equation}
	\lambda_{\text{metabolic}} = 0.005 \cdot \left(\frac{18}{\text{age}}\right)^{1.3}.
\end{equation}
This SOC optimization step is completely decoupled from the neural network training. It is solved using an independent numerical solver leveraging Brent's method (\texttt{scipy.optimize.minimize\_scalar}) \cite{Brent1973, Virtanen2020}, which deterministically scales the \(V_{\text{max}}\) manipulation to find the precise optimal dose for each unique patient.

\section{Hybrid model performance and latent state extraction}

The empirical fMRI dataset utilized for the validation of this model was procured from the open-access Neuroimaging Informatics Tools and Resources Clearinghouse (NITRC). Specifically, the cohort integrates resting-state functional sequences from the ADHD-200 repository \cite{ADHD200}, preprocessed through the Athena pipeline \cite{Bellec2016}, alongside neuroimaging volumes from the Autism Brain Imaging Data Exchange (ABIDE-I) \cite{DiMartino2014}. Together, these data repositories provided a comprehensive, multi-site collection of standardized continuous fMRI arrays (formatted as \texttt{.nii.gz} volumes) required for the subsequent latent feature extraction. 

Following strict quality control protocols requiring exact 400-region parcellation dimensions, the dataset was curated to 2265 valid fMRIs from 1260 unique patients. To strictly prevent data leakage during training, a patient-level GroupKFold split (\(k=5\)) was employed, partitioning the cohort into 1,812 training samples and 453 validation samples per fold.

The model was optimized over 50 epochs per fold. Across the five cross-validation folds, the hybrid architecture achieved a robust average validation accuracy of 75.67\% \(\pm\) 3.62\%. The convergence of the BCE loss combined with the continuous symptom penalty term evidenced significant stability, validating the network's capacity to extract physiologically meaningful latent variables (\(\mathbf{Z}_{\text{global}}\)) directly from the noisy, high-dimensional fMRI arrays.

To visualize the underlying functional connectome utilized by the model, neuroimaging representations were generated using the NiiVue extension for Visual Studio Code. Saliency mapping allowed the extraction of critical spatial features for four specific ROI, alongside a global connectivity mask, which were exported as NIfTI (\texttt{.nii.gz}) volumes. This integration ensures that the mathematical abstractions remain grounded in observable neurobiology.

To clarify the operational distinction within the exported neuroimaging files, the framework differentiates between the global connectivity configurations and the localized saliency mappings. global NIfTI volumes represent a static, temporally integrated overview of the entire functional connectome. Mathematically, this corresponds to the dense vector \(\mathbf{Z}_{\text{global}}\) obtained after the average temporal pooling layer, capturing the entire baseline state of the patient's brain network. Conversely, the saliency NIfTI volumes function as the diagnostic attention maps of the deep learning architecture. Derived directly from the gradient weights of the ViT's self-attention mechanism, these volumes isolate the specific node strengths and regional micro-fluctuations, such as within the fronto-striatal circuits, that the neural network prioritized to classify the disorder and parameterize the subsequent stochastic control equations.

\section{Tables}\label{sec5}

\begin{sidewaystable}
    \caption{Phenotypics, extracted latent parameters and optimization results for the entire validation cohort.}
    \label{tab:combined_all_patients}
    \begin{tabular*}{\textwidth}{@{\extracolsep{\fill}} l c c c c c c c c c c c c @{}}
        \toprule
        & \multicolumn{6}{c}{\textbf{Phenotypics}} & \multicolumn{4}{c}{\textbf{Latent parameters}} & \multicolumn{2}{c}{\textbf{Optimization results}} \\
        \cmidrule{2-7} \cmidrule{8-11} \cmidrule{12-13}
        \textbf{ID} & \textbf{Source} & \textbf{Center} & \textbf{Age} & \textbf{Gender} & \textbf{DX} & \textbf{CM} & \(b_{\text{eff}}\) & \(l_{\text{eff}}\) & \(H_{\text{eff}}\) & \(k_{\text{disc}}\) & \textbf{\(\mathcal{D}^*\) (mg)} & \textbf{\(\mathcal{J}\)} \\
        \midrule
        3235580 & ADHD-200 & NYU  & 8y 7m   & Female & ADHD-C  & anxiety dis. & 0.4509 & 0.1303 & 0.3179 & 0.2212 & 33.22 & 0.8299 \\
        2026113 & ADHD-200 & KKI  & 13y     & Female & ADHD-C  & none & 0.4315 & 0.1357 & 0.3283 & 0.2390 & 56.93 & 0.5870 \\
        1992284 & ADHD-200 & NYU  & 7y 3m   & Male   & ADHD-C  & none & 0.4512 & 0.1345 & 0.3047 & 0.2265 & 20.28 & 0.9202 \\
        2570769 & ADHD-200 & NYU  & 13y 6m  & Male   & ADHD-C  & dyslexia/graphia  & 0.4310 & 0.1051 & 0.3564 & 0.2690 & 57.47 & 0.5613 \\
        3205761 & ADHD-200 & PeU  & 14y 7m  & Male   & ADHD-C  & ODD & 0.3322 & 0.1370 & 0.3581 & 0.3248 & 61.63 & 0.5250 \\
        3286474 & ADHD-200 & OSHU & 8y 1m   & Female & ADHD-HI & none & 0.4106 & 0.1339 & 0.3211 & 0.2796 & 30.54 & 0.8749 \\
        2074737 & ADHD-200 & NI   & 14y 1m  & Male   & ADHD-HI & none & 0.4277 & 0.0934 & 0.3272 & 0.3257 & 59.09 & 0.5389 \\
        1208586 & ADHD-200 & NI   & 20y 11m & Male   & ADHD-HI & none & 0.4273 & 0.1032 & 0.3114 & 0.3189 & 71.11 & 0.3593 \\
        10108   & ADHD-200 & NYU  & 17y 2m  & Female & ADHD-I  & none & 0.4414 & 0.0945 & 0.3618 & 0.2715 & 65.42 & 0.4402 \\
        10106   & ADHD-200 & NYU  & 11y 10m & Male   & ADHD-I  & none & 0.4379 & 0.0938 & 0.3412 & 0.2979 & 51.45 & 0.6348 \\
        3515506 & ADHD-200 & NI & 18y 10m & Male   & ADHD-I  & none & 0.4631 & 0.0950 & 0.3428 & 0.2384 & 67.61 & 0.4007 \\
        50960   & ABIDE    & NYU  & 38y 9m  & Female & ASD     & ADHD-I & 0.3825 & 0.1063 & 0.3485 & 0.3347 & 72.00 & 0.1962 \\
        51328   & ABIDE    & LMU  & 22y     & Male   & ASD     & none & 0.3313 & 0.1282 & 0.3362 & 0.3528 & 72.00 & 0.3414 \\
        51021   & ABIDE    & NYU  & 23y 8m  & Male   & ASD     & ADHD-I            & 0.3537 & 0.1153 & 0.3388 & 0.3514 & 72.00 & 0.3157 \\
        10077   & ADHD-200 & NYU  & 8y      & Female & NT      & none              & 0.3584 & 0.1085 & 0.3461 & 0.3513 & - & - \\
        4256491 & ADHD-200 & PeU  & 12y     & Female & NT      & none              & 0.1297 & 0.2167 & 0.2703 & 0.4359 & - & - \\
        16071   & ADHD-200 & PbU  & 18y 4m  & Female & NT      & none              & 0.1373 & 0.2366 & 0.2757 & 0.4267 & - & - \\
        2601925 & ADHD-200 & KKI  & 9y 7m   & Male   & NT      & none              & 0.1629 & 0.2142 & 0.2818 & 0.4172 & - & - \\
        8692452 & ADHD-200 & NYU  & 12y 3m  & Male   & NT      & none              & 0.2489 & 0.1880 & 0.3058 & 0.3697 & - & - \\
        15057   & ADHD-200 & WU   & 21y 3m  & Male   & NT      & none              & 0.1586 & 0.2047 & 0.2906 & 0.4201 & - & - \\
        \botrule
    \end{tabular*}
\end{sidewaystable}

\begin{sidewaystable}
    \caption{NIfTI visualization thresholds for the whole validation cohort computed in Algorithm \ref{alg:thresholds}.}
    \label{tab:thresholds}
    \begin{tabular*}{\textheight}{@{\extracolsep\fill} l c c c c c c c c c c c c @{}}
        \toprule
        & \multicolumn{2}{@{}c@{}}{\textbf{Global}} & \multicolumn{2}{@{}c@{}}{\textbf{Saliency}} & \multicolumn{2}{@{}c@{}}{\textbf{Attention}} & \multicolumn{2}{@{}c@{}}{\textbf{Motivation}} & \multicolumn{2}{@{}c@{}}{\textbf{Hyperactivity}} & \multicolumn{2}{@{}c@{}}{\textbf{Impulsivity}} \\
        \cmidrule{2-3} \cmidrule{4-5} \cmidrule{6-7} \cmidrule{8-9} \cmidrule{10-11} \cmidrule{12-13}
        \textbf{ID} & \textbf{Min} & \textbf{Max} & \textbf{Min} & \textbf{Max} & \textbf{Min} & \textbf{Max} & \textbf{Min} & \textbf{Max} & \textbf{Min} & \textbf{Max} & \textbf{Min} & \textbf{Max} \\
        \midrule
        10077   & 0.5026 & 0.9900 & 0.5026 & 0.9900 & 0.5570 & 0.6793 & 0.6630 & 0.9900 & 0.5154 & 0.9080 & 0.4433 & 0.8588 \\
        10106   & 0.4391 & 0.8424 & 0.2854 & 0.8424 & 0.1617 & 0.6204 & 0.1650 & 0.8424 & 0.1738 & 0.7898 & 0.1557 & 0.7995 \\
        10108   & 0.4320 & 0.9250 & 0.2951 & 0.9250 & 0.0842 & 0.6623 & 0.0769 & 0.9250 & 0.0897 & 0.7744 & 0.0758 & 0.7538 \\
        15057   & 0.4434 & 0.8541 & 0.4434 & 0.8541 & 0.4982 & 0.6116 & 0.4987 & 0.7585 & 0.4426 & 0.7939 & 0.4692 & 0.8541 \\
        16071   & 0.4230 & 0.7999 & 0.3778 & 0.9219 & 0.4621 & 0.6330 & 0.3773 & 0.7740 & 0.4230 & 0.7218 & 0.4029 & 0.8000 \\
        50960   & 0.5554 & 0.9219 & 0.3778 & 0.9219 & 0.0685 & 0.8841 & 0.0469 & 1.0000 & 0.0476 & 0.9340 & 0.0460 & 0.9219 \\
        51021   & 0.4590 & 0.8475 & 0.3548 & 0.8475 & 0.0547 & 0.8475 & 0.0362 & 1.0000 & 0.0421 & 0.7262 & 0.0458 & 0.8408 \\
        51328   & 0.5136 & 0.9201 & 0.3966 & 0.9201 & 0.0655 & 0.8653 & 0.0503 & 1.0000 & 0.0482 & 0.8295 & 0.0464 & 0.9967 \\
        1208586 & 0.4360 & 0.8593 & 0.2707 & 0.8593 & 0.0485 & 0.6521 & 0.0515 & 0.8593 & 0.0515 & 0.7405 & 0.0433 & 0.7977 \\
        1992284 & 0.4136 & 0.8596 & 0.3555 & 0.8596 & 0.3666 & 0.6618 & 0.2717 & 0.8596 & 0.3555 & 0.7648 & 0.3003 & 0.7101 \\
        2026113 & 0.4963 & 0.8590 & 0.3158 & 0.8590 & 0.1630 & 0.6752 & 0.1472 & 0.8152 & 0.1408 & 0.8524 & 0.1460 & 0.8590 \\
        2074737 & 0.4234 & 0.8976 & 0.2706 & 0.8976 & 0.1252 & 0.5956 & 0.1283 & 0.8976 & 0.1212 & 0.8165 & 0.1011 & 0.7818 \\
        2570769 & 0.4354 & 0.8540 & 0.2721 & 0.8540 & 0.1308 & 0.5963 & 0.1276 & 0.8540 & 0.1357 & 0.8230 & 0.1155 & 0.7313 \\
        2601925 & 0.4673 & 0.8419 & 0.4673 & 0.8419 & 0.5392 & 0.6324 & 0.4870 & 0.7500 & 0.5067 & 0.8076 & 0.5140 & 0.8419 \\
        3205761 & 0.4754 & 0.9171 & 0.3128 & 0.9171 & 0.1143 & 0.6158 & 0.1450 & 0.9171 & 0.1144 & 0.9168 & 0.0994 & 0.7616 \\
        3235580 & 0.4399 & 0.8637 & 0.3372 & 0.8637 & 0.2960 & 0.6642 & 0.2313 & 0.8637 & 0.2883 & 0.8324 & 0.2312 & 0.7057 \\
        3286474 & 0.5017 & 0.9006 & 0.4197 & 0.9006 & 0.4420 & 0.8196 & 0.3647 & 0.8190 & 0.2957 & 1.0000 & 0.3112 & 0.9006 \\
        3515506 & 0.4255 & 0.7983 & 0.2576 & 0.7983 & 0.0677 & 0.5530 & 0.0659 & 0.7847 & 0.0739 & 0.7983 & 0.0541 & 0.8007 \\
        4256491 & 0.4324 & 0.8827 & 0.4324 & 0.8827 & 0.4892 & 0.6818 & 0.3667 & 0.8827 & 0.4423 & 0.7719 & 0.4599 & 0.8006 \\
        8692452 & 0.4248 & 0.7940 & 0.4248 & 0.7940 & 0.4699 & 0.5936 & 0.4031 & 0.7940 & 0.4758 & 0.7470 & 0.4599 & 0.7127 \\
        \botrule
    \end{tabular*}
\end{sidewaystable}

\clearpage

\section{Figures}\label{sec6}


\begin{figure}[!htbp]
    \centering
    \includegraphics[width=0.48\linewidth]{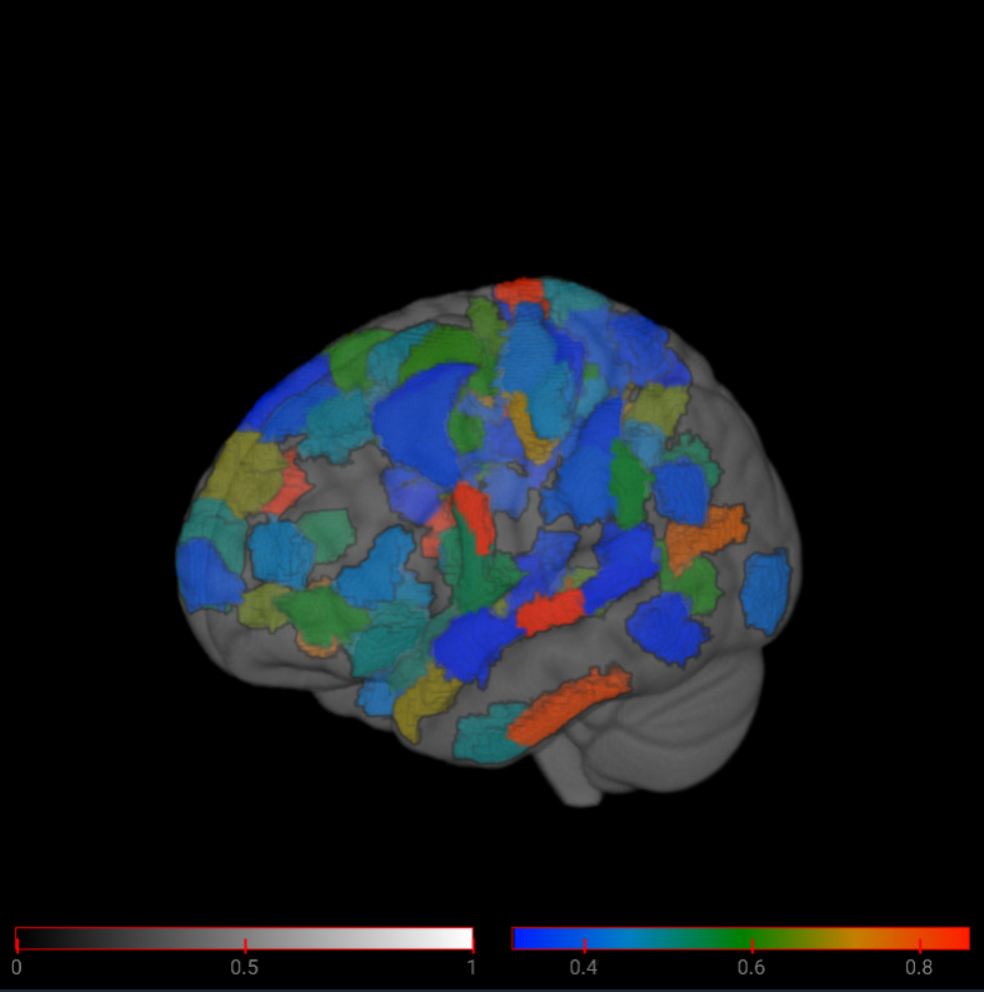}\hfill
    \includegraphics[width=0.48\linewidth]{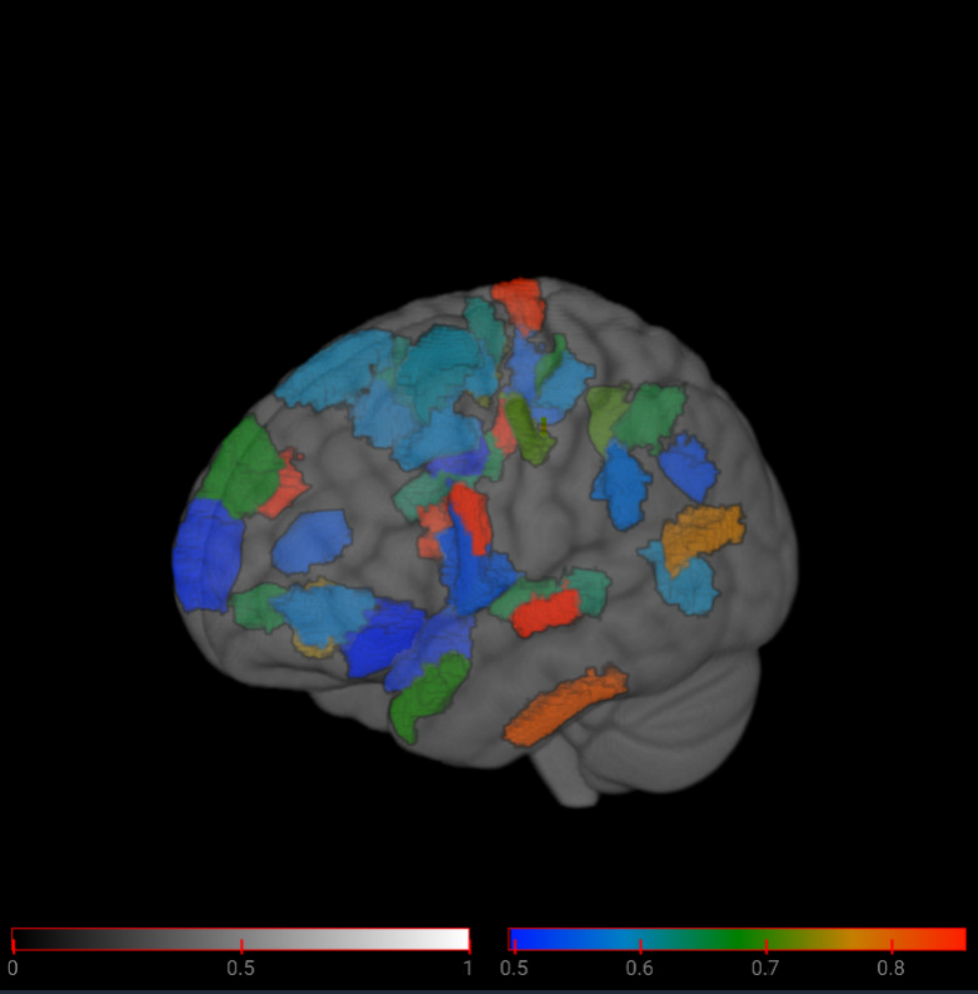}
    \caption{Baseline saliency (left) and global network activations (right) for patient 2026113.}
    \label{fig:2026113_global}
\end{figure}

\begin{figure}[t]
    \centering
    \includegraphics[width=0.48\linewidth, trim={0cm 0cm 0cm 3cm}, clip]{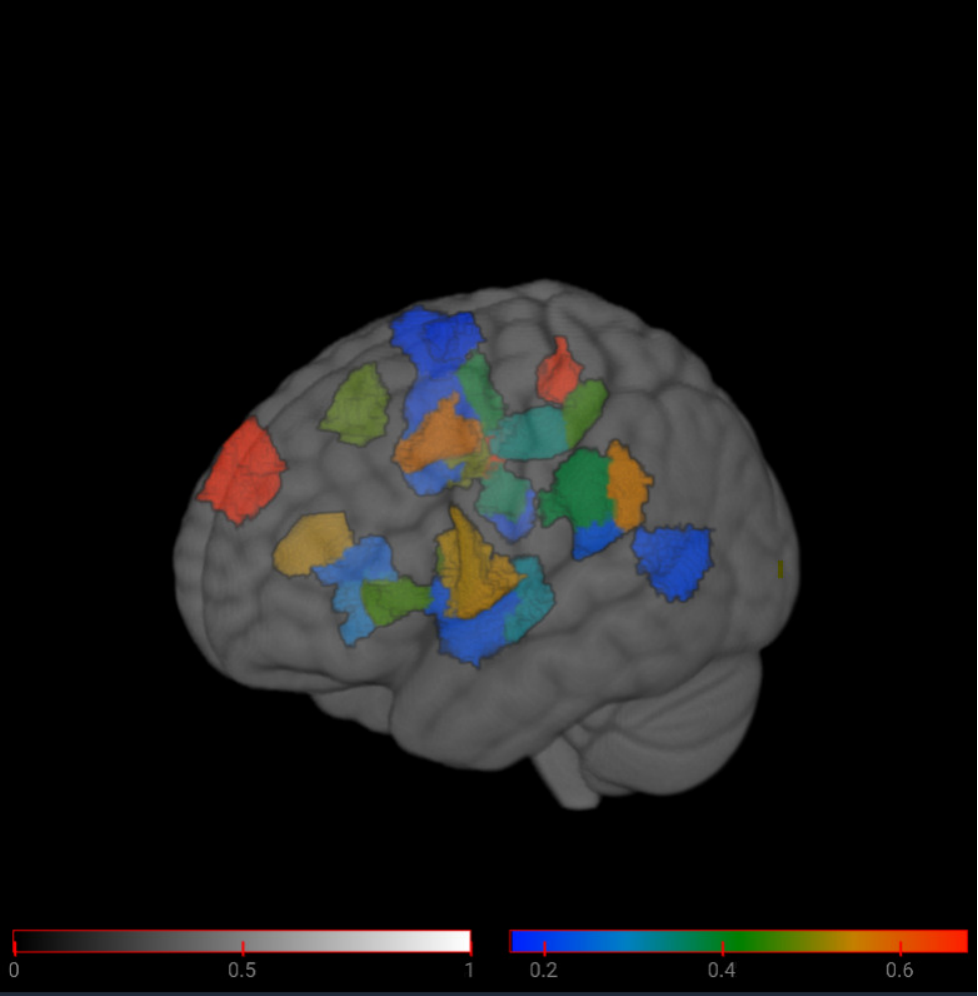}\hfill
    \includegraphics[width=0.48\linewidth, trim={0cm 0cm 0cm 3cm}, clip]{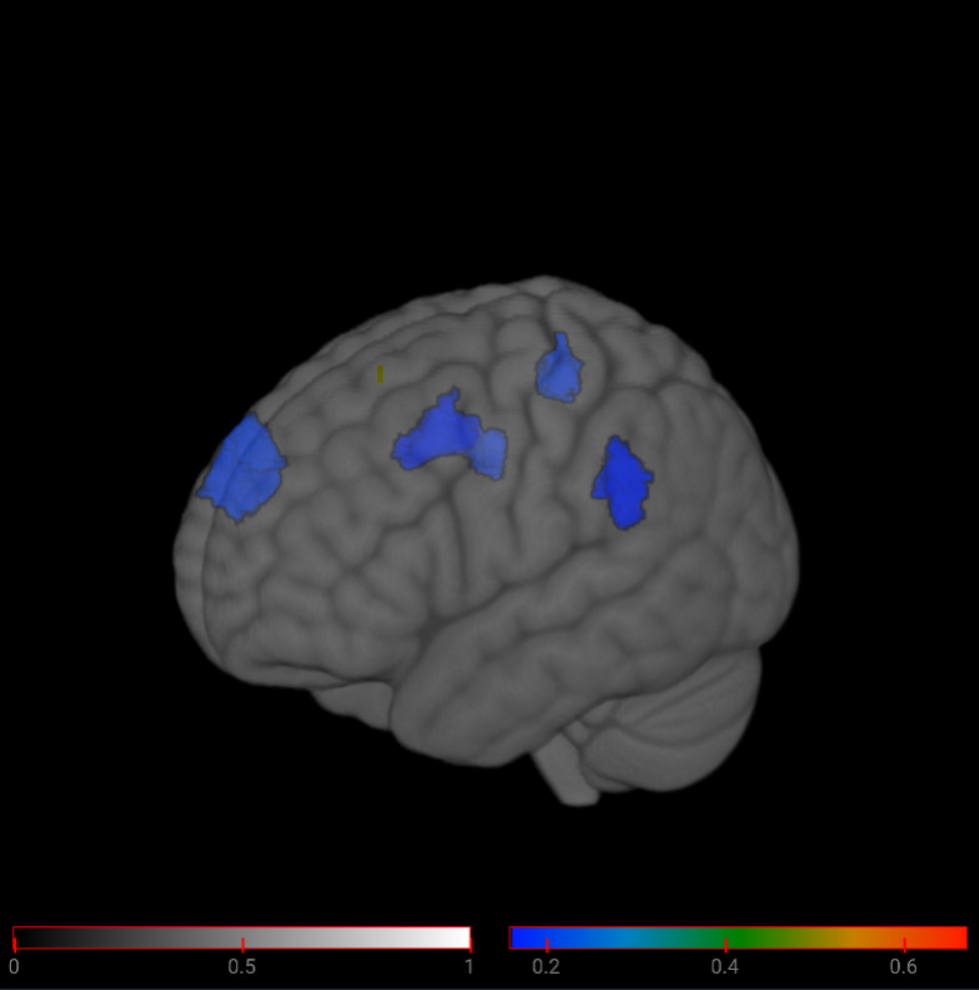}
    
    \vspace{2mm}
    \includegraphics[width=0.48\linewidth, trim={0cm 0cm 0cm 3cm}, clip]{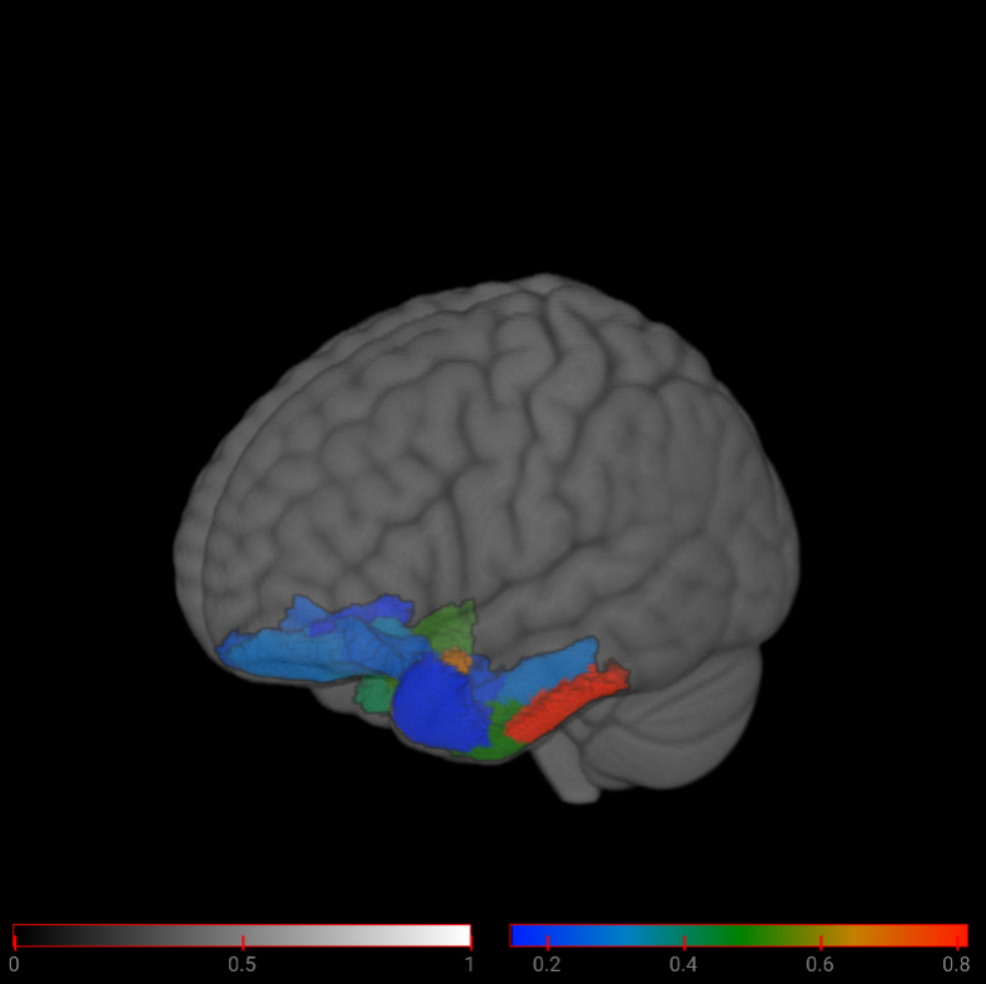}\hfill
    \includegraphics[width=0.48\linewidth, trim={0cm 0cm 0cm 3cm}, clip]{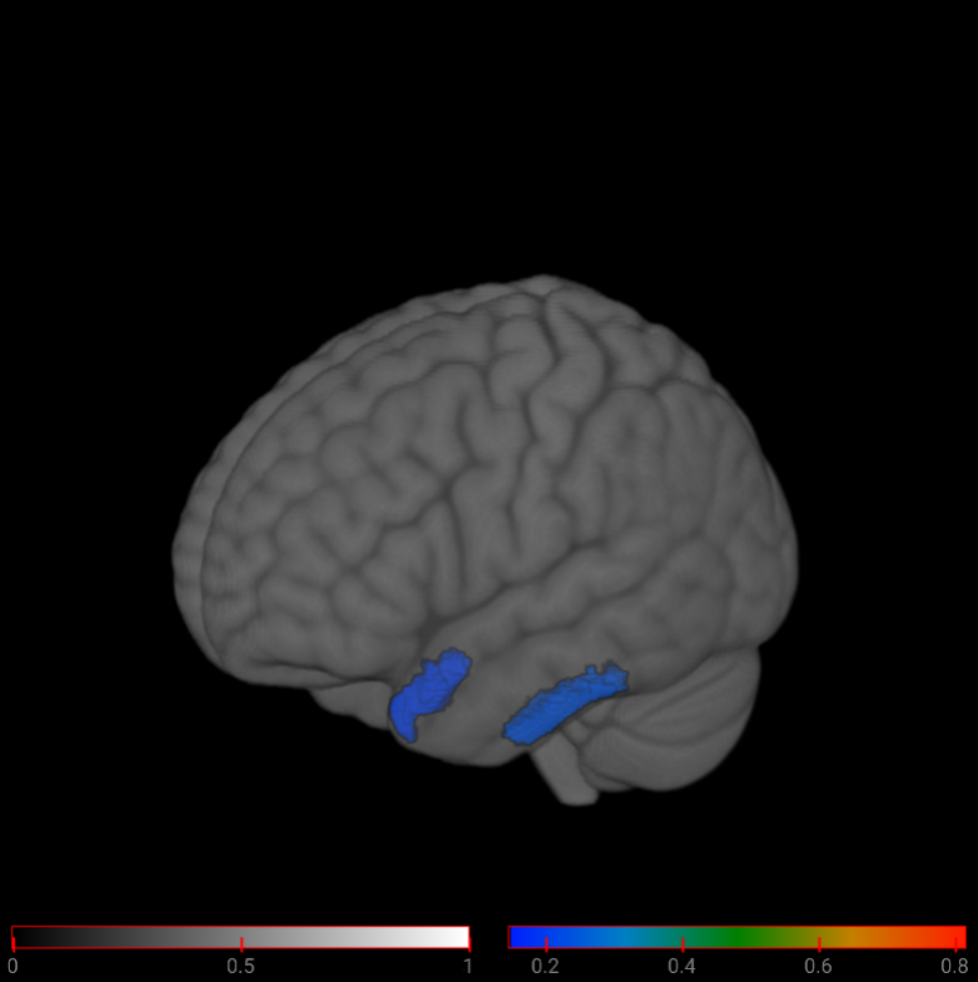}
    
    \vspace{2mm}
    \includegraphics[width=0.48\linewidth, trim={0cm 0cm 0cm 3cm}, clip]{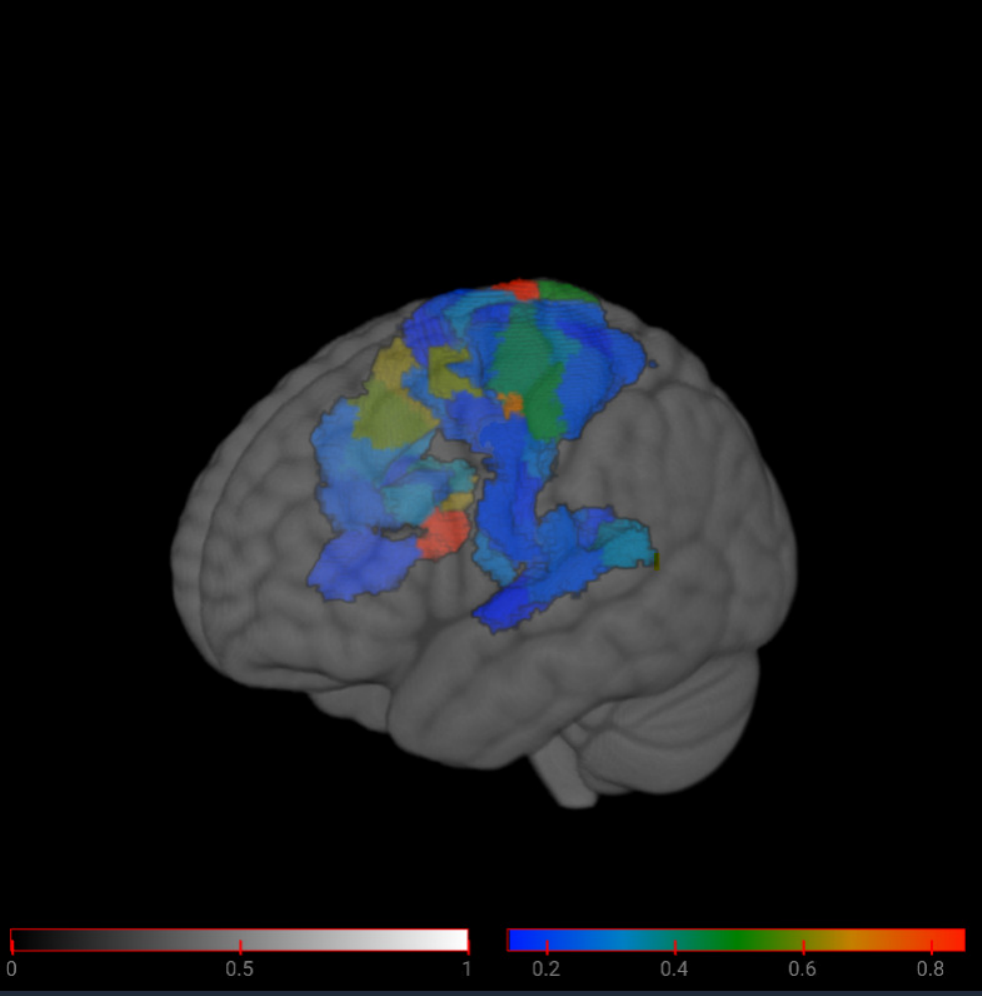}\hfill
    \includegraphics[width=0.48\linewidth, trim={0cm 0cm 0cm 3cm}, clip]{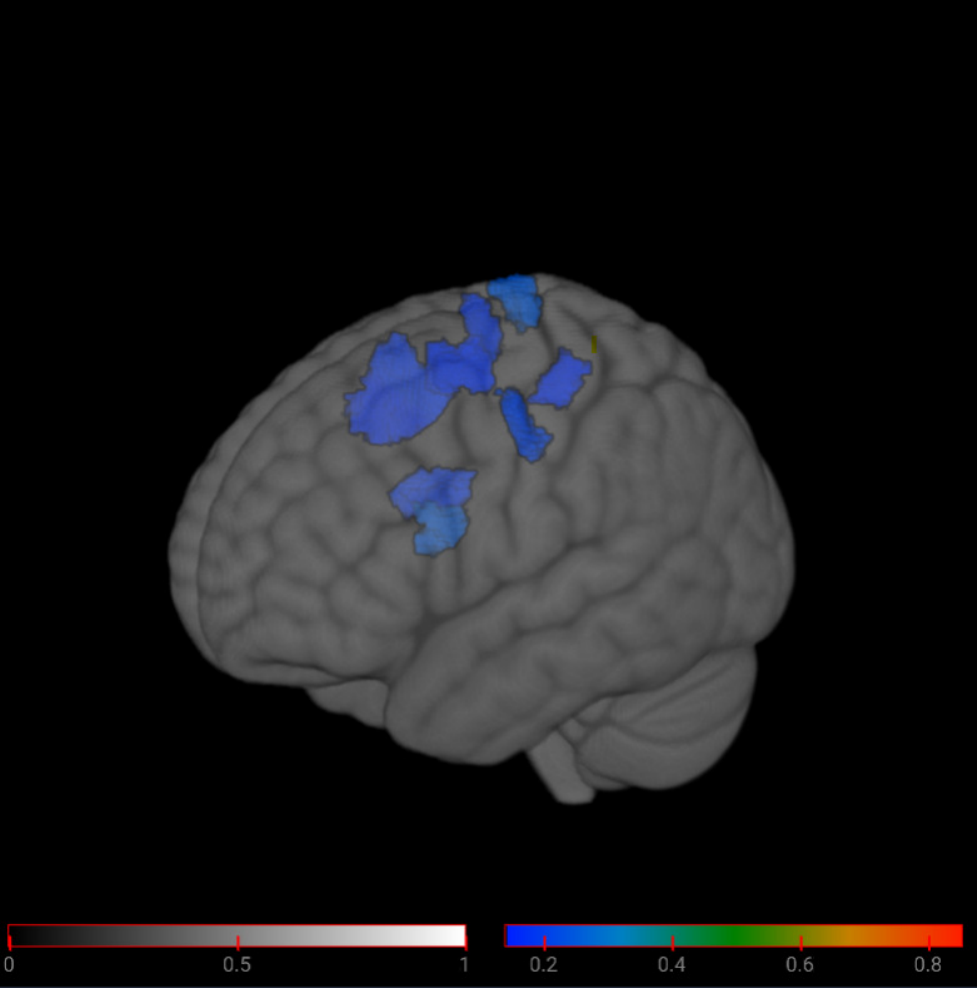}
    
    \vspace{2mm}
    \includegraphics[width=0.48\linewidth, trim={0cm 0cm 0cm 3cm}, clip]{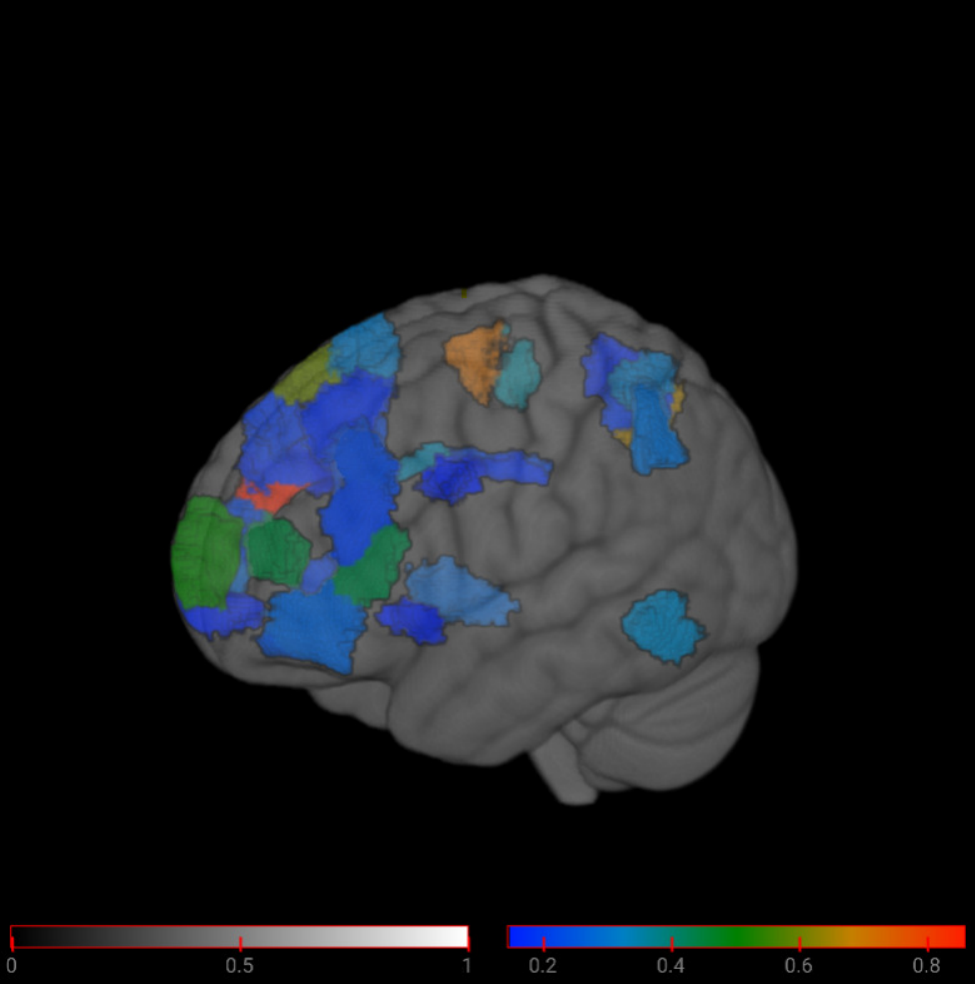}\hfill
    \includegraphics[width=0.48\linewidth, trim={0cm 0cm 0cm 3cm}, clip]{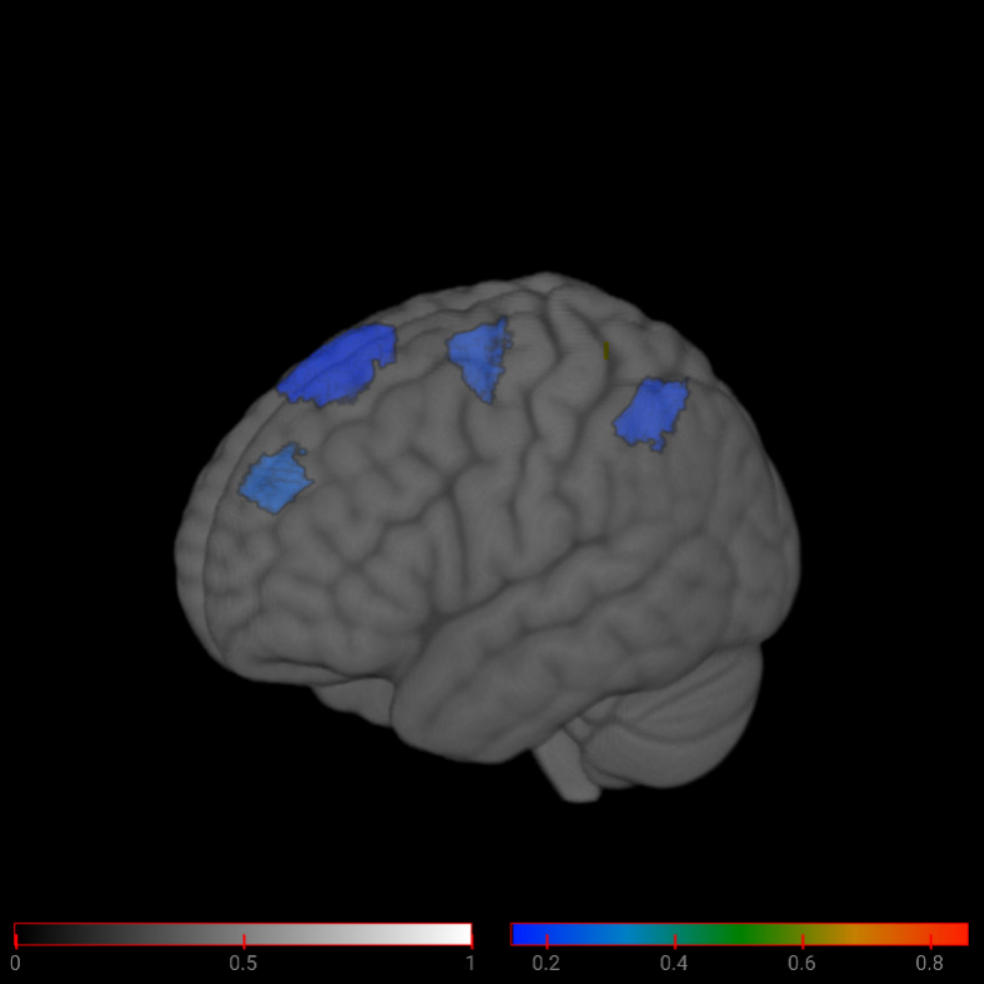}
    
    \caption{Pre and post-intervention spatial activations for patient 2026113, displaying structural normalization under optimal MPH dosage. Rows from top to bottom: attention decay, motivation loss, hyperactivity and impulsivity: left column - baseline state, right column - medicated state.}
    \label{fig:comparative_2026113}
\end{figure}

\clearpage


\begin{figure}[t]
    \centering
    \includegraphics[width=0.48\linewidth]{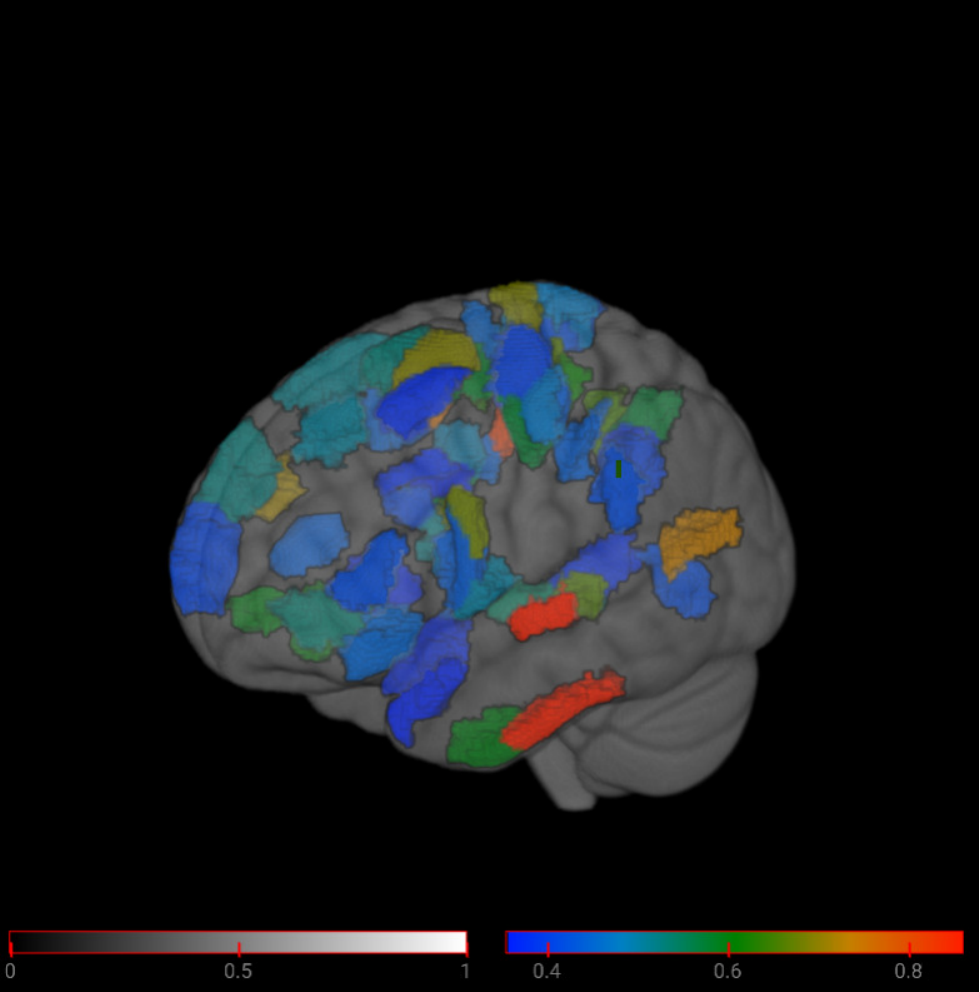}\hfill
    \includegraphics[width=0.48\linewidth]{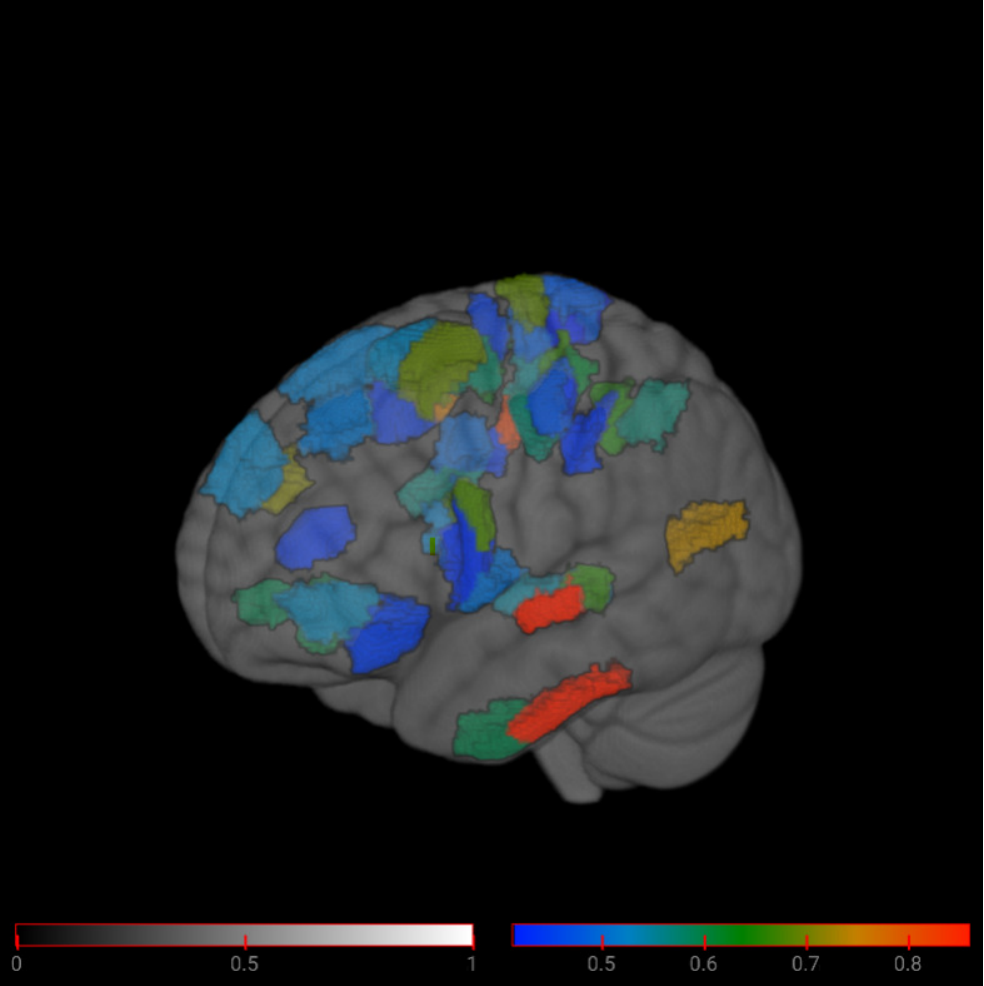}
    \caption{Baseline saliency (left) and global network activations (right) for patient 1992284.}
    \label{fig:1992284_global}
\end{figure}

\begin{figure}[t]
    \centering
    \includegraphics[width=0.48\linewidth, trim={0cm 0cm 0cm 3cm}, clip]{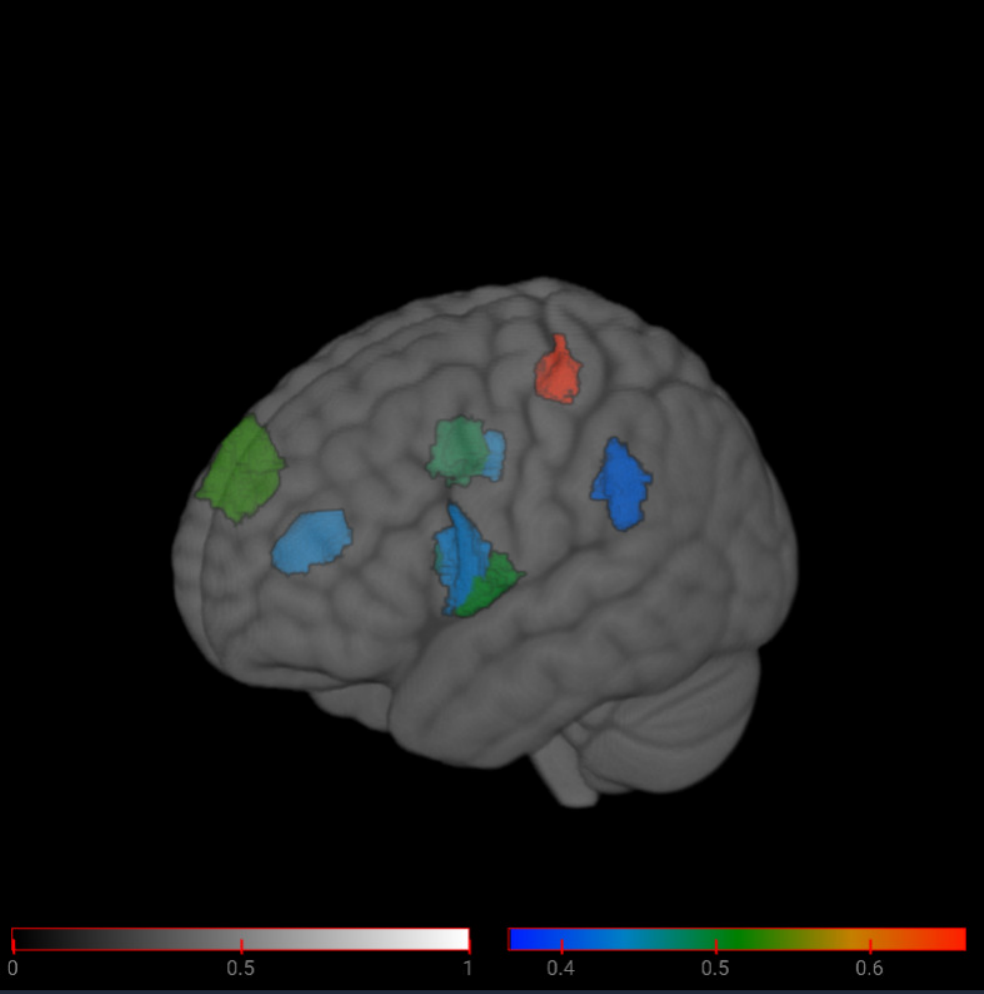}\hfill
    \includegraphics[width=0.48\linewidth, trim={0cm 0cm 0cm 3cm}, clip]{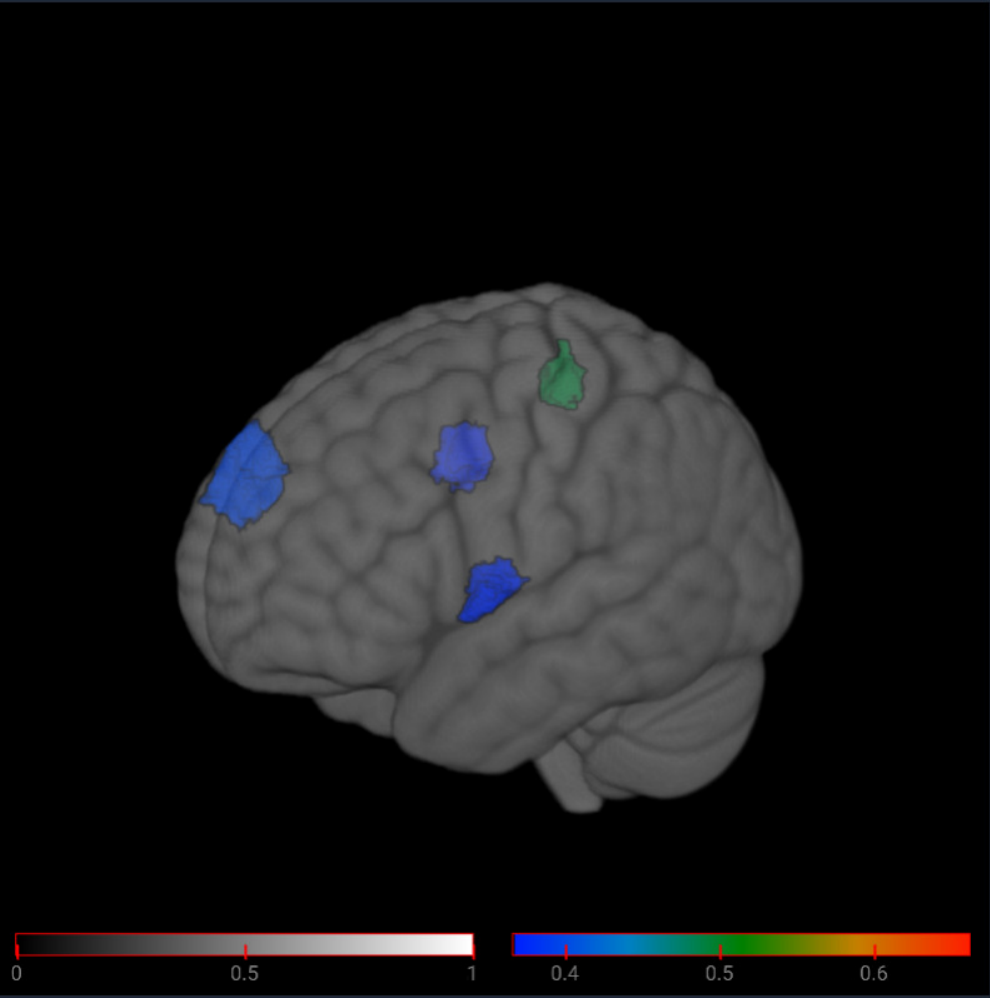}
    
    \vspace{2mm}
    \includegraphics[width=0.48\linewidth, trim={0cm 0cm 0cm 3cm}, clip]{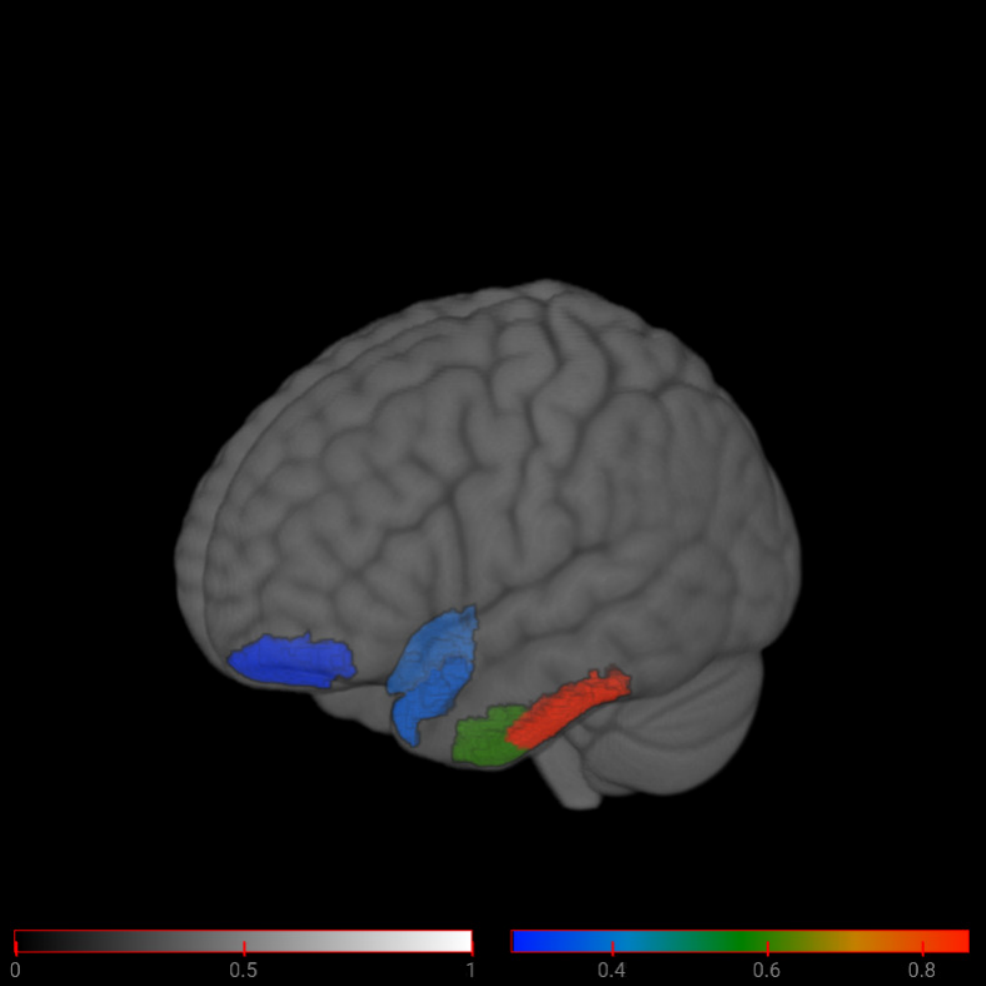}\hfill
    \includegraphics[width=0.48\linewidth, trim={0cm 0cm 0cm 3cm}, clip]{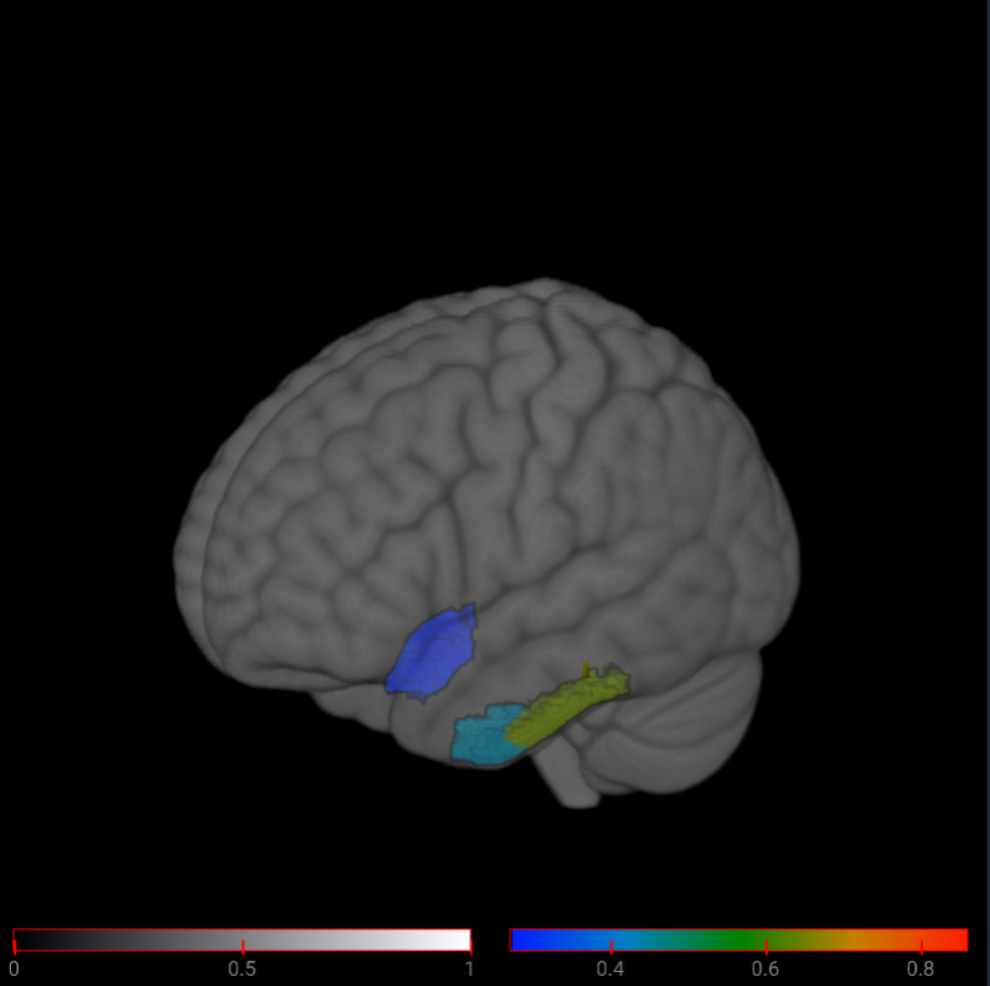}
    
    \vspace{2mm}
    \includegraphics[width=0.48\linewidth, trim={0cm 0cm 0cm 3cm}, clip]{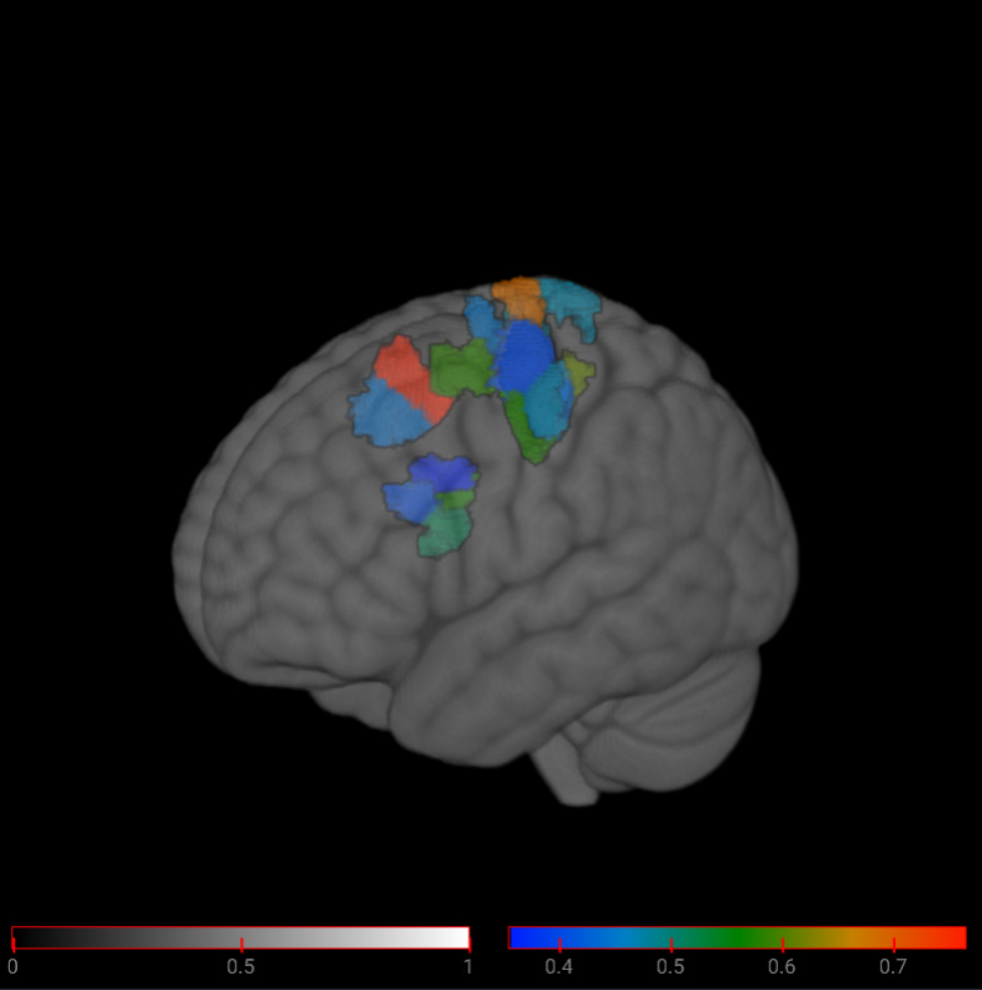}\hfill
    \includegraphics[width=0.48\linewidth, trim={0cm 0cm 0cm 3cm}, clip]{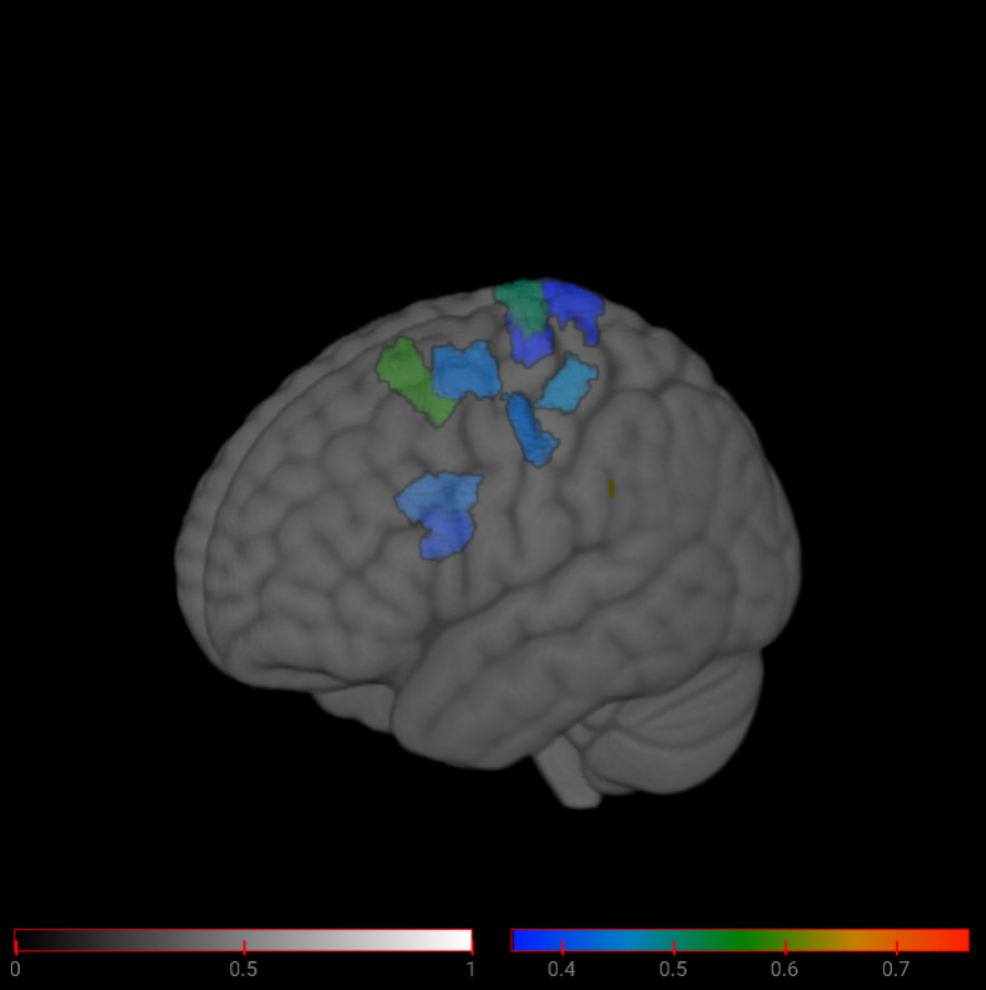}
    
    \vspace{2mm}
    \includegraphics[width=0.48\linewidth, trim={0cm 0cm 0cm 3cm}, clip]{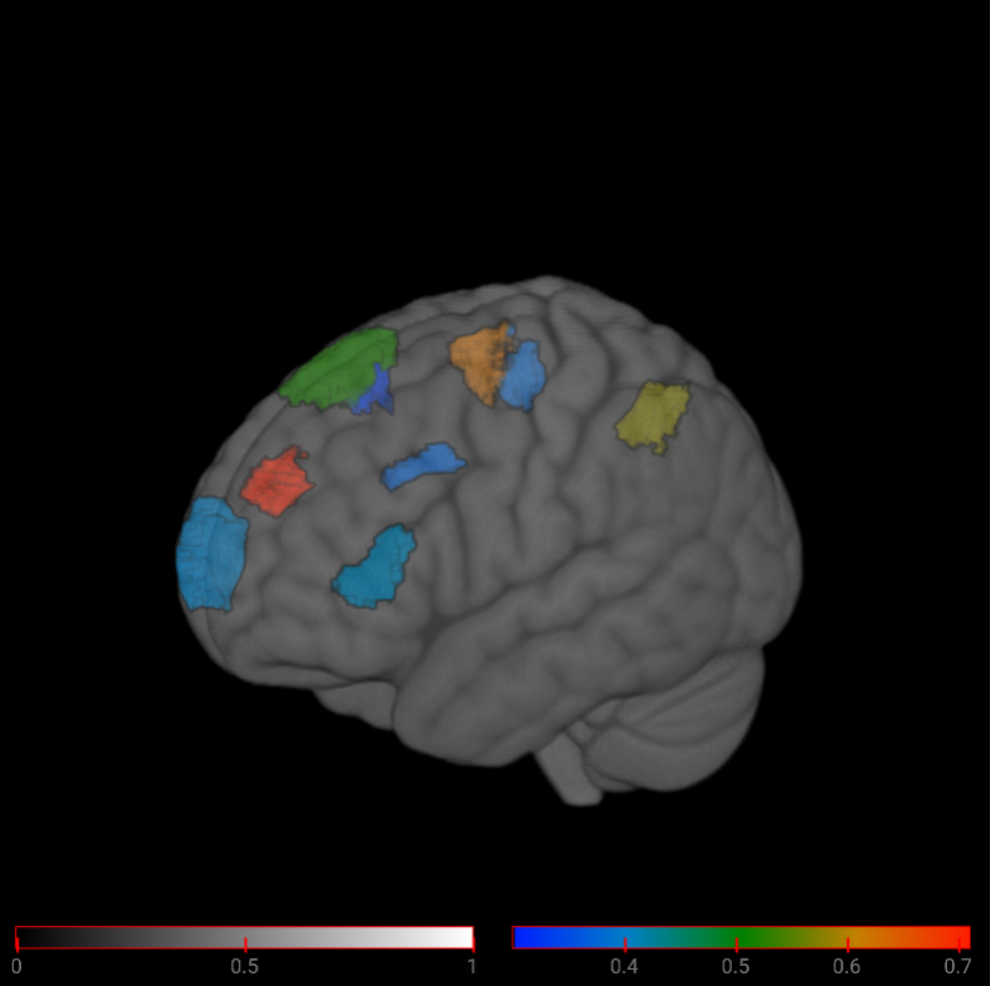}\hfill
    \includegraphics[width=0.48\linewidth, trim={0cm 0cm 0cm 3cm}, clip]{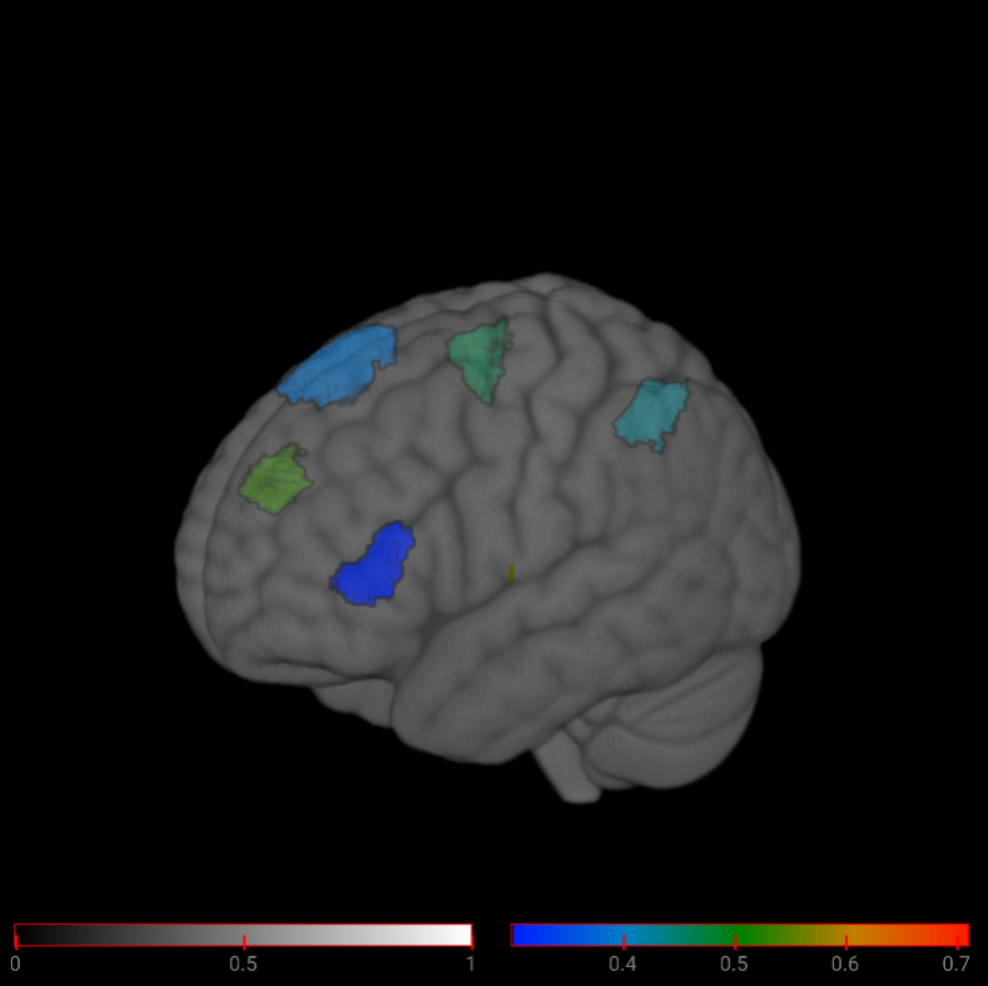}
    
    \caption{Pre- and post-intervention spatial activations for patient 1992284, displaying structural normalization under optimal MPH dosage. Rows from top to bottom: attention decay, motivation loss, hyperactivity and impulsivity: left column - baseline state, right column - medicated state.}
    \label{fig:comparative_1992284}
\end{figure}

\clearpage

\section{Algorithms}\label{sec7}

The computational implementation of the mathematical framework is divided into three primary algorithms. Algorithm \ref{alg:training} describes the deterministic representation learning pipeline (combining the 1D-CNN and ViT architecture with the regularized optimization loop), while Algorithm \ref{alg:optimization} details the Euler-Maruyama integration of the symptom SDEs and the continuous scalar optimization of the external control policy. Furthermore, Algorithm \ref{alg:thresholds} outlines the empirical thresholding procedure utilized to determine standardized visualization boundaries for the continuous NIfTI neuroimaging volumes.

\begin{algorithm}[H]
\caption{Training loop with regularized empirical risk minimization}
\label{alg:training}
\begin{algorithmic}[1]
\State Consider the training dataset of fMRI sequences $\mathbf{X}$ and diagnostic labels $\mathbf{Y}$, number of epochs $N$, penalty weight $\lambda$
\State Initialize hybrid model parameters $\Theta$ (1D-CNN, Positional Embeddings, ViT, MLP)
\For{$epoch = 1, \dots, N$}
    \For{each batch $(\mathbf{x}, y) \in (\mathbf{X}, \mathbf{Y})$}
        \State $\mathbf{f} \gets \text{1D-CNN}(\mathbf{x})$ \# extract local hemodynamic features
        \State $\mathbf{f} \gets \mathbf{f} + \mathbf{P}_{\text{pos}}$ \# inject temporal order 
        
        \State $\mathbf{h} \gets \text{TransformerEncoder}(\mathbf{f})$ \# apply self-attention mechanism
        \State $\mathbf{Z}_{\text{global}} \gets \displaystyle \frac{1}{T} \sum_{t=1}^T \mathbf{h}_t$ \# average temporal pooling for global representation
        
        \State $\hat{y} \gets \text{MLP}(\mathbf{Z}_{\text{global}})$ \# compute diagnostic logits
        \State $\mathcal{L}_{\text{BCE}} \gets \text{BCE}(\hat{y}, y)$ \# calculate maximum likelihood loss
        \State $\mathcal{L}_{\text{penalty}} \gets \text{ComputeSymptomPenalty}(\mathbf{Z}_{\text{global}}, y)$ \# calculate biophysical constraint
        \State $\mathcal{L}_{\text{total}} \gets \mathcal{L}_{\text{BCE}} + \lambda \cdot \mathcal{L}_{\text{penalty}}$ \# compute total functional
        
        \State Update parameters $\Theta$ via backpropagation utilizing $\nabla_{\Theta} \mathcal{L}_{\text{total}}$
    \EndFor
\EndFor
\State \Return Optimized model $\Theta$ mapping high-dimensional fMRI to $\mathbf{Z}_{\text{global}}$
\end{algorithmic}
\end{algorithm}

\begin{algorithm}[H]
\caption{Euler-Maruyama integration and patient-specific optimization}
\label{alg:optimization}
\begin{algorithmic}[1]
\State Consider the global latent representation $\mathbf{Z}_{\text{global}}$, patient age, time domain $[0, T]$ with step $\Delta t$
\State \textbf{\# Step I: parameterize the patient phenotype}
\State $(b_{\text{eff}}, l_{\text{eff}}, H_{\text{eff}}, k_{\text{disc}}) \gets \text{ScaledSigmoid}(\mathbf{Z}_{\text{global}})$ \# deterministic symptom drifts
\State $\sigma \gets \text{StandardDeviation}(\mathbf{Z}_{\text{global}})$ \# biological noise volatility
\State Determine dosage bounds $[\mathcal{D}_{\text{min}}, \mathcal{D}_{\text{max}}]$ according to clinical age guidelines
\State Initialize age-dependent metabolic cost: $\lambda_{\text{metabolic}} \gets 0.005 \cdot \left(\frac{18}{\text{age}}\right)^{1.3}$

\State \textbf{\# Step II: definine the cost functional}
\Function{$\mathcal{J}$}{$\mathcal{D}$}
    \State $\text{PK}(t) \gets \left[ 0.95 \left(1 - e^{-t/0.8}\right) e^{-t/16} \right] \cdot \frac{\mathcal{D}}{54}$ \# instantiate continuous control policy
    \State Initialize symptom arrays $S(0)$ and generate independent Wiener increments $\Delta W \sim \mathcal{N}(0, \Delta t)$
    
    \For{$i = 1, \dots, N_t$} \# Euler-Maruyama integration
        \State Compute exponential decays for attention ($A_t$), motivation ($M_t$) and impulsivity ($I_t$)
        \State $\mu(t) \gets 0.35 + H_{eff} \sin(2\pi \cdot 0.4 \cdot t_{i-1})$ \# set periodic mean target
        \State $H_i \gets H_{i-1} + \theta (\mu(t) - H_{i-1})\Delta t + \sigma \Delta W$ \# Ornstein-Uhlenbeck process for hyperactivity
        
        \State $\mathbf{u}_t^{\text{eff}} \gets \text{PK}(t_{i-1}) \cdot (1.0 - S_{\text{ND}}(t_{i-1}))$ \# apply effective restorative force
        \State Update medicated trajectories using $\mathbf{u}_t^{\text{eff}}$
        \State Apply biological hard-clipping $S \gets \min(\max(S, 0), 1.05)$ to all trajectories
    \EndFor
    
    \State Compute MSE between medicated and unmedicated trajectories (1.0 for A and M, 0.2 for H and 0.1 for I)
    \State \Return $\text{MSE}_{\text{tracking}} + \lambda_{\text{metabolic}} \cdot \mathcal{D}$
\EndFunction

\State \textbf{\# Step III: execute the scalar optimization}
\State $\mathcal{D}^* \gets \text{MinimizeScalar}(\mathcal{J}, \text{bounds}=[\mathcal{D}_{\text{min}}, \mathcal{D}_{\text{max}}], \text{method}=\text{Brent})$
\State \Return Optimal patient-specific dose $\mathcal{D}^*$, simulated baseline and medicated trajectories
\end{algorithmic}
\end{algorithm}

\begin{algorithm}[H]
\caption{Empirical thresholding for NIfTI visualization boundaries}
\label{alg:thresholds}
\begin{algorithmic}[1]
\State Consider the unmedicated and medicated \texttt{.nii.gz} volumes, $\mathbf{X}_{\text{unmed}}$ and $\mathbf{X}_{\text{med}}$, respectively.
\State $\mathbf{D}_{\text{unmed}} \gets \text{LoadData}(\mathbf{X}_{\text{unmed}})$ \# extract unmedicated data array
\State $\mathbf{D}_{\text{opt}} \gets \text{LoadData}(\mathbf{X}_{\text{med}})$ \# extract medicated data array

\State \# Isolate non-zero voxel intensities based on the unmedicated baseline:
\State $\mathbf{Mask} \gets (\mathbf{D}_{\text{unmed}} \neq 0)$
\State $\mathbf{V}_{\text{unmed}} \gets |\mathbf{D}_{\text{unmed}}[\mathbf{Mask}]|$
\State $\mathbf{V}_{\text{opt}} \gets |\mathbf{D}_{\text{opt}}[\mathbf{Mask}]|$

\State $\mathbf{V} \gets \mathbf{V}_{\text{unmed}} \cup \mathbf{V}_{\text{opt}}$ \# combine non-zero absolute intensities

\If{$|\mathbf{V}| > 0$}
    \State $T_{\text{min}} \gets \text{Percentile}(\mathbf{V}, 90)$ \# noise floor filtering
    \State $T_{\text{max}} \gets \text{Percentile}(\mathbf{V}, 99.5)$ \# peak saturation mitigation
    \State \Return $\text{Round}(T_{\text{min}}, 4), \text{Round}(T_{\text{max}}, 4)$
\Else
    \State \Return $0.0, 0.0$
\EndIf
\end{algorithmic}
\end{algorithm}

\section{Conclusion}\label{sec12}

This paper successfully established and validated a novel neurocomputational framework that bridges the theoretical rigor of stochastic measure theory with the empirical demands of clinical psychiatry. By conceptualizing the neurobiological dysfunctions of ADHD as a system of continuous-time stochastic processes, this research moved beyond descriptive statistics, offering a predictive and mathematically optimized paradigm for pharmacological intervention.

The foundational achievement of this work lies in the development of the hybrid 1D-CNN and ViT architecture. Overcoming the limitations of early baseline models, the integration of a one-dimensional convolutional front-end provided the essential temporal inductive bias required to process high-dimensional fMRI sequences. Furthermore, the engineering of the latent chunking mechanism transformed a standard classification task into a precise parameter extraction engine. By enforcing the network to orthogonally encode the four core behavioral axes of ADHD, the spatial attention weights were successfully translated into discrete, clinically relevant scalar parameters.

These extracted parameters served as the critical bridge to the mathematical domain, directly parameterizing the SDEs governing each patient's symptomatology. Modeling executive and motivational deficits as exponential decays and motor hyperactivity as an Ornstein-Uhlenbeck process provided a mathematical representation of the unmedicated ADHD baseline. This stochastic formalization accurately captured the inherent tendency of the ND connectome to converge towards stable, yet dysfunctional behavioral regimes under the influence of physiologic Brownian noise.

Ultimately, the OROS formulation of the ER MPH PDs as a continuous control policy represented the pinnacle of the framework. By solving the SOC problem, the mathematical optimizer successfully computed patient-specific dosages, being capable of counteracting the ND-specific dysfunctional drifts. Interpreted probabilistically as a change of measure, the control signal effectively guided the ND-specific trajectories towards NT targets. The validation of these numerical outcomes against the spatial attention maps confirmed that our mathematical optimization reliably mirrors the structural normalization of the fronto-striatal networks, establishing a computational proof of concept.

\backmatter

\newpage

\section{List of abbreviations}

\begin{table}[h!]
    \caption{List of abbreviations}\label{tab:abbreviations}
    \begin{tabular}{@{} p{3cm} l @{}}
        \toprule
        \textbf{Abbreviation} & \textbf{Definition} \\
        \midrule
        \textbf{1D-CNN} & One-dimensional convolutional neural network \\
        \textbf{ACC} & Anterior cingulate cortex \\
        \textbf{ADHD} & Attention-deficit/hyperactivity disorder \\
        \textbf{ADHD-C} & Combined presentation of ADHD \\
        \textbf{ADHD-HI} & Predominantly hyperactive/impulsive presentation of ADHD \\
        \textbf{ADHD-I} & Predominantly inattentive presentation of ADHD \\
        \textbf{A-S} & Asperger's syndrome \\
        \textbf{ASD} & Autism spectrum disorder \\
        \textbf{BCE} & Binary cross-entropy \\
        \textbf{BOLD} & Blood-oxygen-level-dependent \\
        \textbf{CEN} & Central executive network \\
        \textbf{CM} & Comorbidity \\
        \textbf{DAT} & Dopamine transporter \\
        \textbf{DDT} & Dynamic developmental theory \\
        \textbf{dlPFC} & Dorsolateral prefrontal cortex \\
        \textbf{DMN} & Default mode network \\
        \textbf{DSM} & Diagnostic and Statistical Manual of Mental Disorders \\
        \textbf{DX} & Diagnosis \\
        \textbf{ELBO} & Evidence lower bound \\
        \textbf{ER} & Extended-release \\
        \textbf{fMRI} & functional magnetic resonance imaging \\
        \textbf{FPN} & Frontoparietal Network \\
        \textbf{GABA} & \(\gamma\)-aminobutyric acid \\
        \textbf{IR} & Immediate-release \\
        \textbf{KL} & Kullback-Leibler \\
        \textbf{MLE} & Maximum likelihood estimation \\
        \textbf{MNI} & Montreal Neurological Institute \\
        \textbf{MPH} & Methylphenidate \\
        \textbf{MSDL} & Multi-subject dictionary learning \\
        \textbf{MSE} & Mean squared error \\
        \textbf{ND} & Neurodivergent \\
        \textbf{NET} & Norepinephrine transporter \\
        \textbf{NT} & Neurotypical \\
        \textbf{OFC} & Orbitofrontal cortex \\
        \textbf{OROS} & Osmotic-controlled release oral delivery system \\
        \textbf{OU} & Ornstein-Uhlenbeck \\
        \textbf{PD} & Pharmacodynamics \\
        \textbf{PET} & Positron emission tomography \\
        \textbf{PK} & Pharmacokinetics \\
        \textbf{ROI} & Regions of interest \\
        \textbf{SMA} & Supplementary motor area \\
        \textbf{SOC} & Stochastic optimal control \\
        \textbf{SSM} & State-space model \\
        \textbf{TDC} & Typically developing children \\
        \textbf{ViT} & Vision Transformer \\
        \botrule
    \end{tabular}
\end{table}

\clearpage


\end{document}